\documentclass[pdflatex,sn-mathphys-num]{sn-jnl}

\usepackage{graphicx}%
\usepackage{multirow}%
\usepackage{amsmath,amssymb,amsfonts}%
\usepackage{amsthm}%
\usepackage{mathrsfs}%
\usepackage[title]{appendix}%
\usepackage{textcomp}%
\usepackage{manyfoot}%
\usepackage{booktabs}%
\usepackage{algorithm}%
\usepackage{algorithmicx}%
\usepackage[noend]{algpseudocode}
\usepackage{listings}%
\usepackage{caption}
\usepackage{subcaption}
\usepackage{comment}
\usepackage{tablefootnote}
\usepackage{xcolor}
\usepackage{pict2e}
\usepackage{tabularx}
\newsavebox{\ORCIDlogo}
\savebox{\ORCIDlogo}{%
\setlength{\unitlength}{\dimexpr 1em/256\relax}%
\begin{picture}(256,256)%
  \color[HTML]{A6CE39}\put(128,128){\circle*{256}}%
  \color{white}%
  \put(78.6,199.2){\circle*{20}}%
  \moveto(70.9,176,9)\lineto(86.3,176,9)\lineto(86.3,69.8)\lineto(70.9,69.8)%
  \closepath\fillpath%
  \moveto(108.9,176.9)\lineto(150.5,176.9)%
  \curveto(190.1,176.9)(207.5,148.6)(207.5 ,123.3)%
  \curveto(207.5,95,8)(186,69.7)(150.7,69.7)%
  \lineto(108.9,69.7)%
  \closepath\fillpath%
  \color[HTML]{A6CE39}%
  \moveto(124.3,83.6)\lineto(148.8,83.6)%
  \curveto(183.7,83.6)(191.7,110.1)(191.7,123.3)%
  \curveto(191.7,144.8)(178,163)(148,163)%
  \lineto(124.3,163)%
  \closepath\fillpath%
\end{picture}%
}
\newcommand\orcidicon[1]{\href{https://orcid.org/#1}{\usebox{\ORCIDlogo}}}

\begin{document}

\title[Article Title]{Evaluation of Adversarial Robustness in Arabic Language Models}



\author*[1,2]{\fnm{Anwar} \sur{Alajmi}}\email{anwaaralajme@gmail.com}

\author[1]{\fnm{Ayed} \sur{Salman}}\email{ayed.salman@ku.edu.kw}

\author[1]{\fnm{Imtiaz} \sur{Ahmad}}\email{imtiaz.ahmad@ku.edu.kw}

\affil[1]{\orgdiv{Department of Computer Engineering}, \orgname{Kuwait University}, \country{Kuwait}}

\affil[2]{\orgdiv{Department of Computer Science}, \orgname{College of Business Studies, Public Authority of Applied Education and Training}, \country{Kuwait}}


\abstract{The emergence of the recent outstanding capabilities of Arabic Language Models has opened doors for exposing their vulnerabilities. One of the major security risks associated with such Natural Language Processing models is adversarial attacks. These attacks can deceive the model into the wrong prediction, raising critical model security and safety concerns. This study aims to assess the robustness of five state-of-the-art Arabic Language Models under a distinct set of Arabic adversarial attacks applied at various levels of granularity and using different example generation strategies. We also explore a defense technique based on adversarial training to enhance model robustness. The results show that insertion of diacritics can reduce the accuracy of some models by 92\% while maintaining a low perturbation distance. For word-level attacks, manipulating Arabic conjunctions preserves high semantic similarity scores, low perturbation distance, and leads to an accuracy degradation of up to 58\%. For sentence-level attacks, paraphrasing proves its effectiveness by an average reduction of 76\% percent in the victim models’ performance. While adversarial training improves overall resilience, with MARBERT being the most robust and AraBERT showing the greatest relative gains, challenges persist, particularly against character-level noise. These findings highlight both the potential and limitations of current defense strategies in morphologically rich languages like Arabic.}


\maketitle

\newpage
\section{Introduction}
Natural Language Processing (NLP) models achieved remarkable success in a wide variety of applications that integrate with vital aspects of human life including education, medicine, and media. However, recent attacks on these models exposed serious safety concerns. Language Models (LMs) susceptibility to such attacks have critical implications that include the security of the model, the misuse of the compromised models to harm others, leakage of private or sensitive data, as well as economic and social effects \cite{DBLP:journals/corr/AmodeiOSCSM16}.
Attacks that exploit the vulnerabilities of machine learning (ML) models are referred to as adversarial attacks. Adversaries craft malicious adversarial examples through input modifications to intentionally lead the model into a false prediction \cite{cubuk2017intriguing}. 

Textual adversarial attacks on NLP models were introduced through a Computer Vision (CV) attack called Fast Gradient Sign Method (FGSM) \cite{goodfellow2014explaining}. Due to feature space differences, CV adversarial attacks generate poor adversarial samples that don't satisfy the attack constraints such as grammatical correctness and semantic similarity in NLP. Hence, NLP-specific attack strategies were introduced throughout the past years to further examine the weaknesses of LMs \cite{nlprobustnesss}.

\begin{figure}[h]
    \centering
    \includegraphics[width=\linewidth]{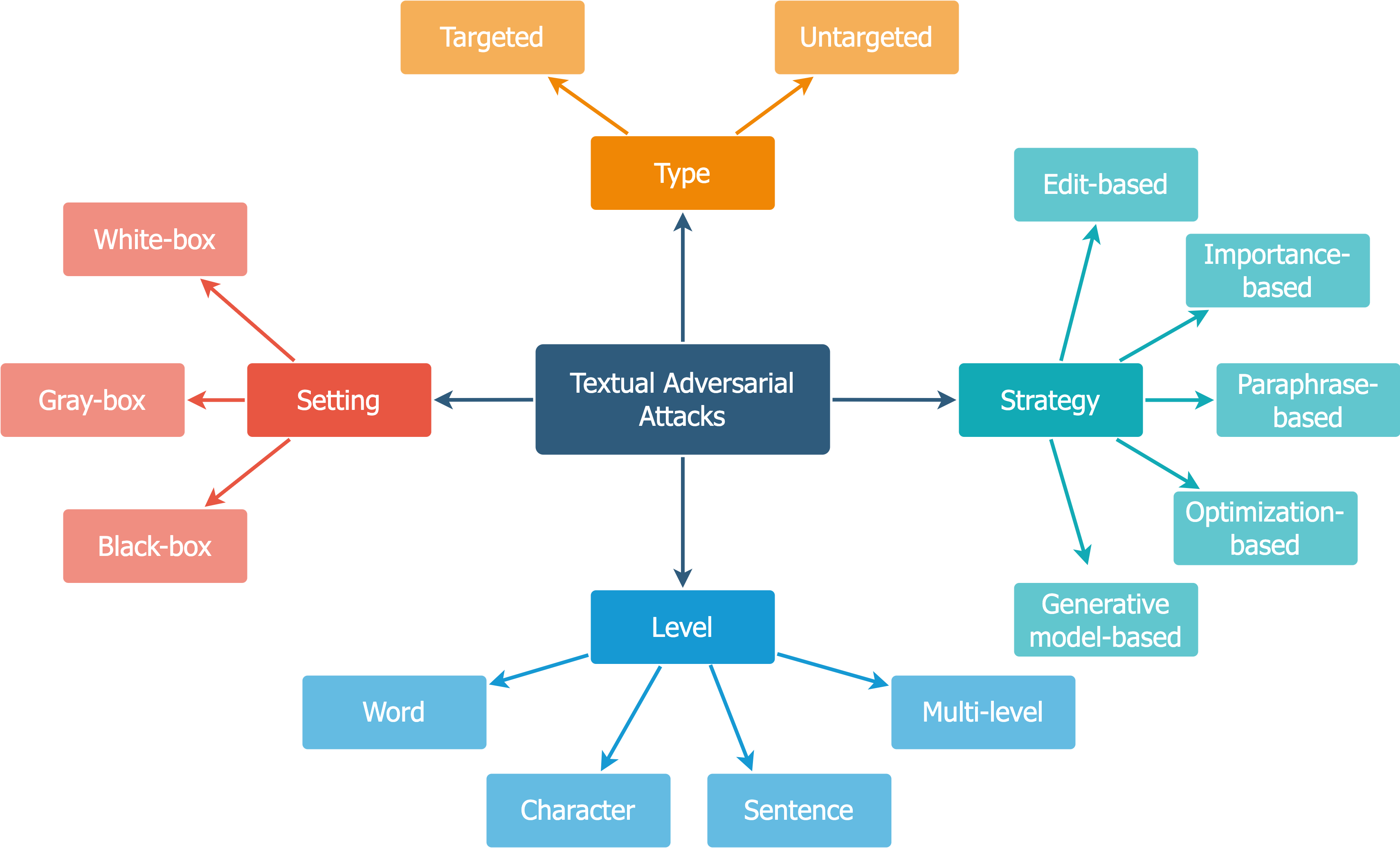}
    \caption{NLP Attacks Taxonomy.}
    \label{Taxonomy}
\end{figure}

Textual adversarial attacks are commonly categorized based on the modified object \cite{ALSHEMALI2020105210}. Character-level attacks occur when a character is inserted, removed, or replaced. Word/token-level attacks happen when the same operations are performed on words/tokens, whereas sentence-level attacks are generated through paraphrasing, substituting, removing, or injecting sentences in the input sequence. Sometimes attacks composed of two or more of the previous granularity levels are referred to as multi-level attacks \cite{nlprobustnesss}. Further categorization is based on the adversarial example-generation strategies, which include edit-based, and importance-based methods \cite{QIU2022278}. In the edit-based techniques, no features or attributes of the modified objects are considered. Meanwhile, importance-based methods usually focus on perturbing object with the most importance to the prediction of the victim model. Figure \ref{Taxonomy} provides a taxonomy of textual adversarial attacks.

Adversarial attacks raise critical ethical concerns, particularly when LMs are deployed in critical applications such as healthcare and legal systems. The exploitation of these models vulnerabilities can lead to data theft, business disruption, misinformation, or harm. The attacks can be designed based on the adversary's full  access to the victim model's parameters (white-box) or without any model-specific knowledge (black-box) \cite{wiyatno2019adversarial}. Usually such attacks occur in the real-world under black-box theme \cite{nlprobustnesss,ALSHEMALI2020105210}. Moreover, targeted attacks occur when the adversary expects a specific misprediction, whereas non-targeted attacks happen when the adversary aims for a general incorrect output \cite{wiyatno2019adversarial,tsipras2018robustness, acloser}. Adversarial robustness evaluation must be a crucial measure to consider before the release of any NLP model to mitigate these issues. While this field gained significant attention among researches worldwide, it is still very understudied when it comes to Arabic models. 
 
 The Arabic language is the 5th most spoken language globally, with over 334 million speakers, making it one of the most popular Semitic languages \cite{ethnologue2025}. Only few limited studies discuss this area due to Arabic language-specific reasons such as low resources, the complexity of the language, and its various dialects \cite{anwar}. Arabic words are complex in structure where a single word can convey the meaning, tense, number, and gender. Furthermore, different vowels (diacritics) on an Arabic word could completely change its meaning \cite{habash2022introduction,al2006design}. Hence, from a computational linguistics and NLP perspective, Arabic presents unique challenges compared to other languages, particularly in areas like adversarial robustness, which require comprehensive evaluations across diverse linguistic nuances.

 Despite significant advancements in NLP, Arabic remains underrepresented when it comes to adversarial robustness research due to its linguistic complexity, dialectical diversity, and resource limitations. These challenges have left a critical gap in evaluating how state-of-the-art Arabic and multilingual NLP models perform under adversarial conditions. Therefore, this study addresses this gap by exploring the effects of non-targeted, black-box adversarial attacks on leading models such as AraBERT \cite{arabert}, MARBERT \cite{marbert}, CaMeLBERT \cite{camel}, mBERT \cite{devlin2018bert}, and XLM-T \cite{xlmt}. The key contributions of this paper are listed as follows:
 \begin{itemize}
     \item This work is the first of its kind to evaluate the adversarial robustness of Arabic NLP models, providing novel insights into their performance under adversarial conditions.
     \item The application of a diverse set of Arabic adversarial attacks on all levels of granularity (character-level, word-level, and sentence-level) using various example-generation strategies (edit-based, importance-based, and paraphrase-based).
     \item A thorough evaluation process that includes measures of accuracy, attack success rate, perturbation rate, and semantic similarity.
 \end{itemize}

The paper is organized as follows: Section \ref{sec2} reviews related studies in the field of LMs adversarial robustness. Sections \ref{sec3} explores the methodology, while Section \ref{sec4} demonstrates and discusses the experimental results. Finally, the conclusion and future work are presented in Section \ref{sec5}.

\section{Literature Review}\label{sec2}

Adversarial attacks are carefully crafted input examples designed to intentionally manipulate NLP models into producing incorrect outputs. Some of these perturbed examples may not be easily detectable by humans but have the potential to deceive LMs. Textual attacks are usually categorized based on the semantic granularity of the perturbed object (character, word, token, sentence) \cite{morris2020reevaluating}. The study by \cite{qiu2019review} further classifies the attacks based on the strategy employed for generating examples. Semantic granularity categories have example generation strategies as sub-categories: gradient-based, optimization-based, importance-based, paraphrase-based, and generative model-based (GAN). Certain criteria must be met when generating adversarial examples that includes grammatical errors that are equal to or less than the original example as well as semantic similarity \cite{chen2021multi}. Numerous methods exist to check the previous requirements such as grammar checking tools, and semantic similarity measures (cosine similarity, universal sentence encoders (USE) \cite{cer2018universal}, and Sentence-BERT \cite{reimers-gurevych-2019-sentence}).

Attacks on the character level involve perturbation of characters that includes adding, removing, swapping, or substituting characters within a given sequence. While these strategies are notably efficient and hard to observe, their detection can be straightforward using spell-checking tool \cite{shreya2022survey}. 
Character-level attacks under black-box theme focus on adding natural or synthetic noise such as punctuation, diacritics, special characters, and numbers. 

While attack methods proposed in \cite{zero, formento-etal-2023-using} are edit-based, the works by \cite{chai2023additive,alshemali2021character} depend on importance-based character transformations. The authors in \cite{chai2023additive}
leverage local post-hoc explanation, SHapley Additive exPlanations (SHAP) \cite{shap} for importance-based adversarial example generation. On the other hand, \cite{alshemali2021character} uses a Word Importance Rank (WIR) to find the most important words for character manipulation.  Moreover, DeepWordBug and TextBugger \cite{gao2018black,li2018textbugger} obtain the importance of the words within a sequence through greedy WIR algorithm, then apply multiple character-level perturbations to the most important word. On the other hand, TextAttack \cite{morris2020textattack} offers a broader framework where character-level manipulations are a part of its toolkit. It also proposes word-level and semantic-level transformations, including paraphrasing and other higher-level alterations.

For word-level attacks, \cite{li-etal-2020-bert-attack, bae, alshemali2019adversarial} employ WIR to craft black-box word replacements. While \cite{li-etal-2020-bert-attack} replaces the word with the highest importance by its synonym, \cite{bae} uses BERT Masked Language Model (MLM) to predict the synonym. On the other hand, \cite{alshemali2019adversarial} perturbs words to violate Arabic grammatical rules in order to deceive victim models including BERT, CNN, and LSTM. In a similar approach to \cite{bae}, \cite{ekbal2022adversarial} uses BERT MLM to replace words with the highest scores. However, instead of conventional scoring functions, SHAP is used to find the sensitivity of the model's prediction to each word within the original sequence, which is similar to the word-attacks in \cite{alajmi2025evaluating}. The work in \cite{alshalan2023attacking} evaluates the robustness of AraBERT and CaMeLBERT under edit-based and gradient-based word substitutions. AraBERT's accuracy decreased by 7\% (edit-based) and 37\% (gradient-based), while CaMeLBERT's accuracy was reduced by 10\% and 36\% after edit-based and gradient-based attacks, respectively. The previously reviewed attack studies are summarized in Table \ref{Robustness Works 1}.

LMs adversarial robustness is typically assessed by evaluating their performance in the face of carefully crafted attacks. It is a measure of how well a model performs under adversarial attacks \cite{ALSHEMALI2020105210,nlprobustnesss}.
To improve the ability of NLP models to resist adversarial attacks, researchers have proposed data-driven approaches that rely on data augmentation. This means training the models on a mixture of real data and synthetic data that has been modified in a way to create adversarial examples. This helps the models to learn to be less sensitive to small changes in the input data, which can make them more robust to adversarial attacks. The work in
\cite{yoo2021towards} utilized data augmentation with less cost. The study presents a simple vanilla adversarial training process (A2T) on BERT and RoBERTa to reduce the computational time and complexity of generating attacks and therefore reduce overall adversarial training time. In general, the approach consists of generating white-box gradient-based attacks under DistilBERT \cite{distilbert} similarity constraint.

Another active defense mechanism is to use a ML attack detector to identify and remove adversarial examples from the training data as proposed by \cite{shen2023textdefense,huber2022detecting}. In \cite{shen2023textdefense} a defense method against importance-based attacks based on deletion-based scoring is proposed. The method calculates the words importance for the original and adversarial sequences. Then, entropy is calculated for both to identify an adversarial example. However, \cite{huber2022detecting} uses SHAP for word importance calculations, and then TextAttack  to craft word-level adversarial attacks. After that, a detection model is trained on the generated examples. Similarly, \cite{mozes2020frequency} calculates the word frequency in the adversarial example and original example to detect word-level perturbations in text classification models. On the other hand, \cite{zhou2019learning} proposes a method to discriminate and restore perturbed tokens semantics through searching the corpus token embedding space using k-nearest neighbors (kNN).

In addition to adversarial training, representation learning is also considered to be an effective and active defense mechanism. This involves using techniques such as randomizing inputs, and unifying input representation to create more robust representations of the input data. An example of this is Dirichlet Neighborhood Ensemble approach by \cite{zhou2020defense}. The authors focus on improving adversarial robustness through exposing the target model to broader forms of the input samples including possible synonyms. This is done through forming virtual sentences by randomized sampling within the convex hull that is spanned by the sequence word embeddings and its synonyms. The process is done for every word within the original input sequence. Then, the model is trained on the augmented data with preserved semantics. Finally, there are a number of passive defense tactics that can be used to protect NLP models from adversarial attacks, such as misspelling check tools used in \cite{gao2018black,li2018textbugger} to defend against character-level attacks \cite{qiu2019review}. A list of the defense methods reviewed is shown in Table \ref{tab:Defense}.

Authors in \cite{alshahrani2024arabic} apply word-level synonym replacement on WordCNN \cite{hochreiter1997long}, and WordLSTM \cite{kim-2014-convolutional}, and AraBERTBASE. Most important words are identified using a ranking greedy algorithm and then top k replacement words are generated from AraBERTBASE v024 as MLM model. WordCNN and WordLSTM accuracy decreased by almost 5\% on MSDA dataset, whereas AraBERT's accuracy decreased by  26.93\%.

The introduction of transformers \cite{vaswani2017attention}, have revolutionized the NLP field through the attention mechanism, which allowed the models to enhance the text context understanding by processing different parts of the input simultaneously. One of the most successful transformer-based models is BERT or Bidirectional Encoder Representations from Transformers which leverages contextualization by its bidirectional input processing approach \cite{DBLP:journals/corr/abs-1810-04805}. Hence, this study only focuses evaluating BERT-based models to examine the capabilities of compromising such powerful models. Since NLP adversarial robustness is limited in the Arabic literature, this study is the first to evaluate a variety of customized Arabic attacks on all levels of granularity on Arabic and multilingual models. Furthermore, it provides insights on the success of the attacks by incorporating a variety of evaluation metrics for measuring the perturbation amount and the semantic similarity of the generated adversarial examples.

\begin{sidewaystable}
\begin{center}
\begin{minipage}{\textheight}
\begin{tabular*}{\textwidth}{@{\extracolsep\fill}ccccp{3cm}p{6cm}}

\toprule
      Reference & Year & Attack level & Strategy & Victim Model & Method \\
          \midrule

    \cite{gao2018black} & 2018 & Character & Importance-based & CNN & Character insertion, deletion, swapping, and replacement\\

    \cite{li2018textbugger} & 2018 & Character & Importance-based & CNN and LSTM & Insertion, deletion, swapping, and replacement\\

    \cite{alshemali2019adversarial} & 2019 & Word & Importance-based & Bi-LSTM, CNN, XLNet, and BERT & Word replacement with Arabic adjectives perturbations \\

     \cite{li-etal-2020-bert-attack} & 2020 & Word & Importance-based & BERT & BERT MLM word substitution \\

    \cite{bae} & 2020 & Word & Importance-based & BERT & BERT MLM word insertion and substiution \\
    
    \cite{zero} & 2020 & Character & Edit-based & RoBERTa & 10 attacks based on character deletion, swapping, and substitution\\
    
      \cite{alshemali2021character} & 2021 &  Character & Importance-based  & BERT, Bi-LSTM, XLNet, and CNN & Substituting Arabic characters with visually similar ones\\

    \cite{ekbal2022adversarial} & 2022 & Word & Importance-based & BERT & BERT MLM word substitution based on SHAP score \\
  
    \cite{formento-etal-2023-using} & 2023 & Character & Edit-based &  BERT, RoBERTa \cite{liu2019roberta}, and DistilBERT \cite{sanh2019distilbert} & Punctuation insertion\\

    \cite{chai2023additive} & 2023 & Character & Importance-based & GRU, LSTM, CNN, and BERT  & Character replacement through LIME \cite{ribeiro2016should} and SHAP \cite{shap}\\

    \cite{alshahrani2024arabic} & 2024 & Word & Importance-based & LSTM, CNN, and AraBERT & Synonym replacement through AraBERT MLM\\

\bottomrule
\end{tabular*}
\caption{Literature review summary of black-box adversarial attacks on NLP models.}
    \label{Robustness Works 1}
\end{minipage}
\end{center}
\end{sidewaystable}

\begin{sidewaystable}
\begin{center}
\begin{minipage}{\textheight}
\begin{tabular*}{\textwidth}{@{\extracolsep\fill}ccclp{6cm}}

 \toprule
      Reference & Year & Attack level  & Victim Model & Method \\
          \midrule
          \cite{gao2018black} & 2018 & Character-level & CNN & Check misspelling to avoid character-level perturbations through Python autocorrect package \\
 
 \cite{li2018textbugger} & 2018 & Character-level & CNN and LSTM & Misspelling checking through a context-aware service \\ 
 
\cite{zhou2019learning} & 2019 & Word-level & BERT & Proposes a KNN based embedding estimator to recover the semantics of perturbed words\\ 
 
\cite{mozes2020frequency} & 2020 & Word-level & CNN, LSTM, and RoBERTa & Using word frequency to detect adversarial sequences examples \\

   \cite{yoo2021towards} & 2021 & Word-level from \cite{morris2020reevaluating} & BERT and RoBERTa & Enhanced adversarial training technique \\
   
   \cite{shen2023textdefense} & 2023 & Word-level & BERT and LSTM & Detection of adversarial attacks through calculating the entropy between the original sample and the adversarial sample \\
   
   \cite{huber2022detecting} & 2022 & Word-level from \cite{morris2020textattack} & Bi-LSTM & Train Bi-LSTM in order to detect word-level attacks \\
 
   \cite{zhou2020defense} & 2020 & Word-level & CNN, LSTM, and bag-of-words & Representation learning through randomized word synonyms substitutions \\
    \bottomrule
    \end{tabular*}
    \caption{Reviewed works on adversarial robustness under black-box attacks.}
    \label{tab:Defense}
\end{minipage}
\end{center}
\end{sidewaystable}

\section{Methodology}\label{sec3}

This section discusses the victim models, the task, and the dataset for fine-tuning the models. In addition, the adversarial setting as well as the character-level, word-level, and sentence-level attacks will be explored.

\subsection{Victim Models}
Due to the state-of-the-art performance of BERT on NLP tasks, specifically for text classification, the adversarial robustness of the most successful Arabic and multilingual BERT-based models will be examined. Hence, the victim models are as follows:
\begin{itemize}
    \item \textbf{AraBERT}: Pre-trained on a large Modern Standard Arabic (MSA) corpus, AraBERT is one of the first BERT models dedicated for Arabic NLP tasks. Due to the complexities within the Arabic language as well as its variant dialects, AraBERT incorporates customized pre-processing and tokenization techniques to comprehend Arabic text \cite{arabert}.

    \item \textbf{MARBERT}:  While AraBERT's pre-training data mainly consist of MSA, MARBERT \cite{marbert} extends AraBERT's approach through including dialectal Arabic. 

    \item \textbf{CaMeLBERT}: Based on BERT architecture, CaMeLBERT (mix) \cite{camel} extends AraBERT and MARBERT by pre-training on large datasets that consist of three types of Arabic text: classical Arabic, MSA, and dialectal Arabic.

    \item \textbf{mBERT}: Extended from BERT, mBERT or Multilingual BERT is pre-trained on a large multilingual corpus of 104 languages including the Arabic language.

    \item \textbf{XLM-T}: XLMs or Cross-lingual Language Models \cite{xlm} differ from BERT in their enhanced capabilities by transferring learning between languages. Hence, these type of models work very well on multiple languages, including low resource languages like Arabic. XLM-T \cite{xlmt}, which is a variant that is trained on Twitter data was selected due to the nature of the used dataset samples.

\end{itemize}

\subsection{Task and Dataset}
The selected task for generating and evaluating the adversarial attacks is binary text-classification. In specific, classical sentiment analysis is selected due to its broad and common uses, without delving into more specific tasks that can cause the victim models to differ in their performances.

The dataset used \cite{Alajmi} is diverse as it is constructed from the Arabic sentiment tweets datasets in \cite{elmadany2018arsas,barbieri-etal-2022-xlm}, which combine tweets in MSA and dialectal Arabic. In the pre-processing phase, unnecessary special characters, including numbers, non-Arabic characters, punctuation, and emojis, are neglected. Note that Arabic diacritics (tashkeel) were not present in the raw dataset and are not affected by this step.”. In addition, Farasa segmentation \cite{abdelali-etal-2016-farasa} is applied to break down Arabic words into stems, prefixes, and suffixes, which will improve performance by reducing redundancy. Finally, the processed text is tokenized using SentencePiece \cite{kudo-richardson-2018-sentencepiece}.

\begin{figure}[h]
\begin{subfigure}{0.45\textwidth}
    \includegraphics[width=0.9\linewidth]{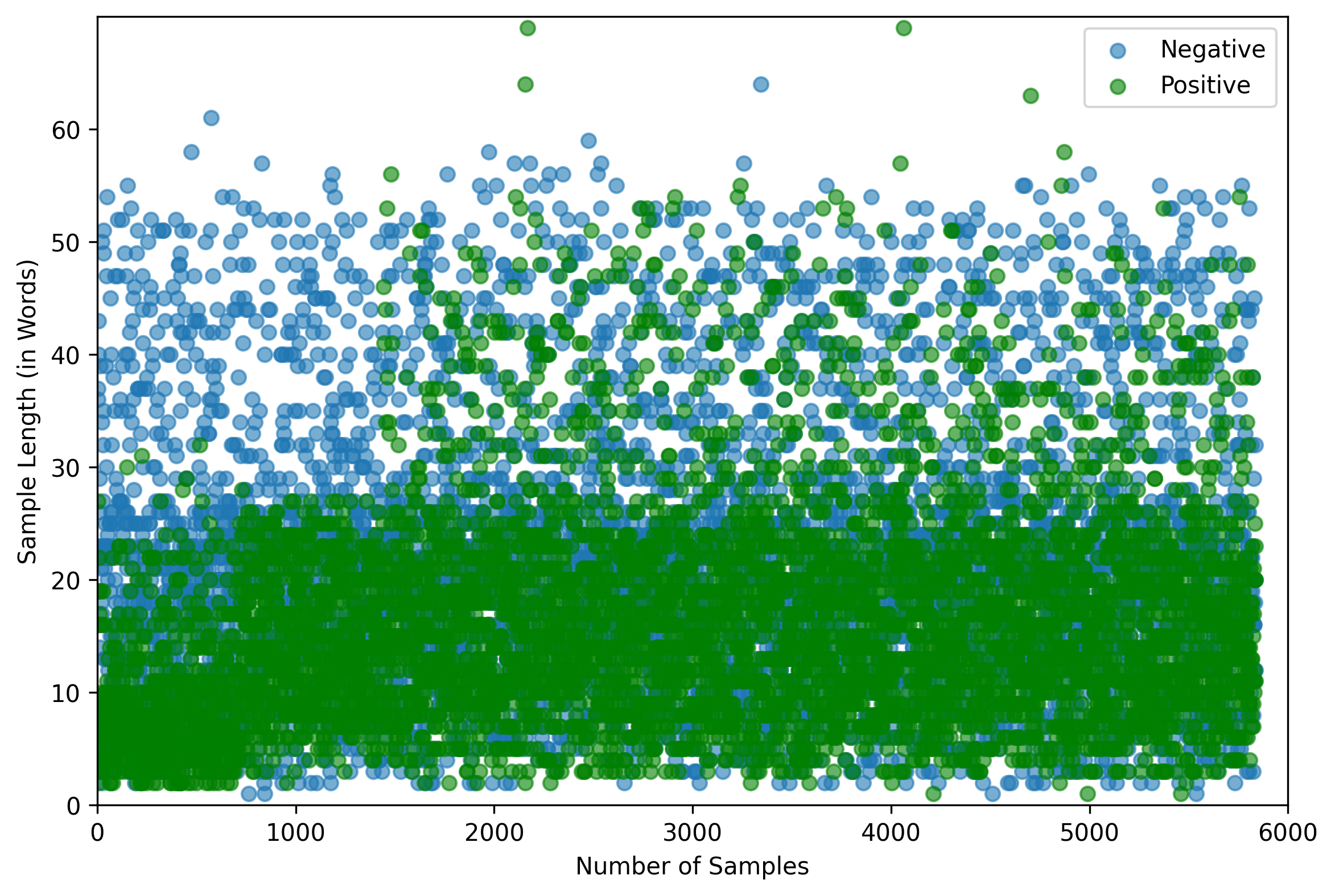} 
    \caption{Classes statistics within the dataset.}
    \label{fig:DSclasses}
\end{subfigure}
\hfill 
\begin{subfigure}{0.45\textwidth}
    \includegraphics[width=0.9\linewidth]{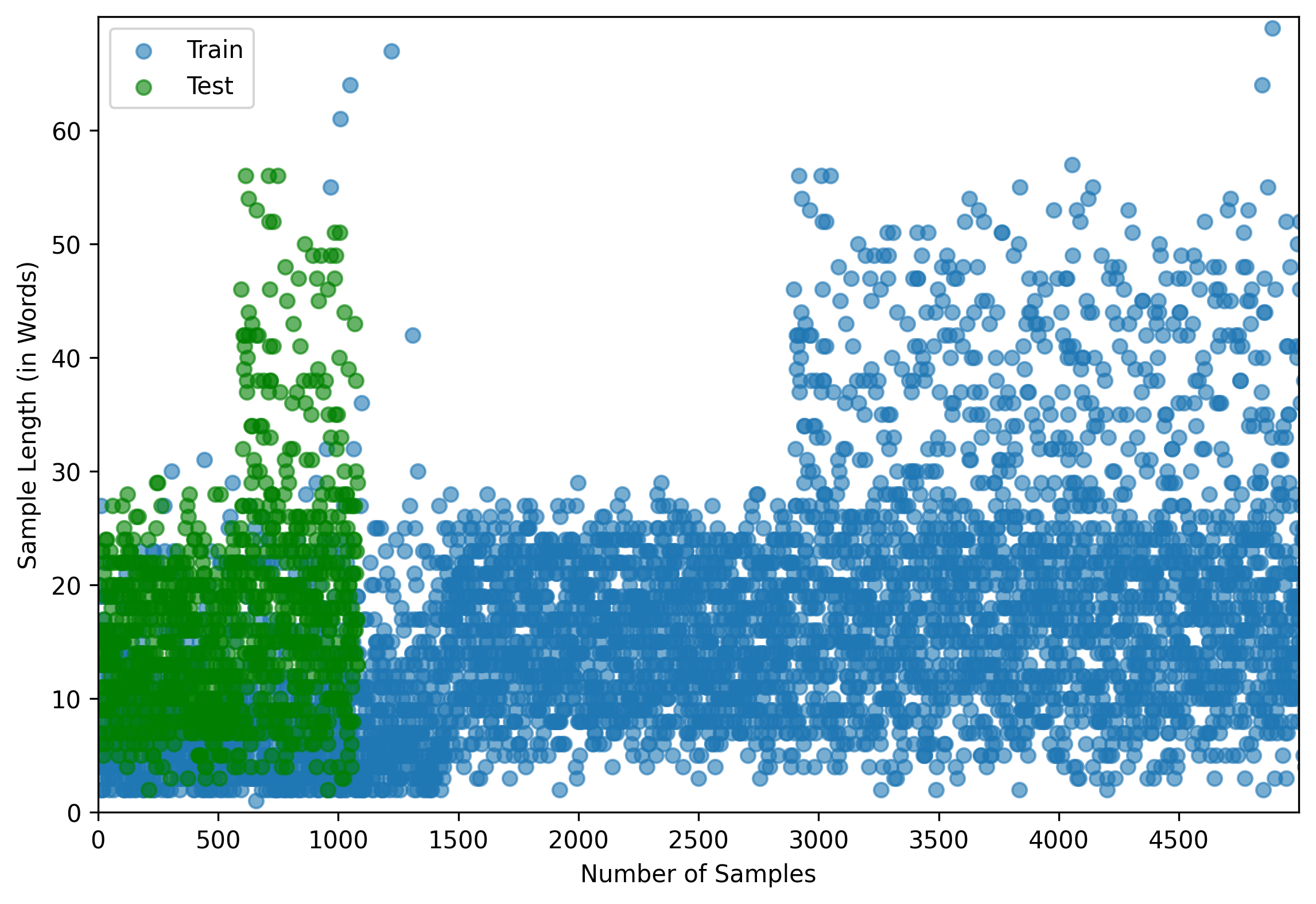}
    \caption{Training and test subsets statistics.}
    \label{fig:DSsets}
\end{subfigure}
\caption{The Used Sentiment Analysis Dataset Information.}
\end{figure}

The dataset includes roughly 16K samples in total are labeled as negative or positive due to the nature of the task (binary). The training set includes 14.7K samples, while the testing set has 1.08K samples. Due to performance issues associated with the imbalanced number of classes, sampling techniques such as down-sampling and up-sampling were carried out to ensure the accuracy and objectivity of the models' prediction. The distribution and length of the sample classes, as well as the training and test sets, are shown in Figures \ref{fig:DSclasses} and \ref{fig:DSsets}. 

\subsection{Adversarial Setting and Objective}
\begin{figure}[h!]
    \centering
    \includegraphics[width=0.7\linewidth]{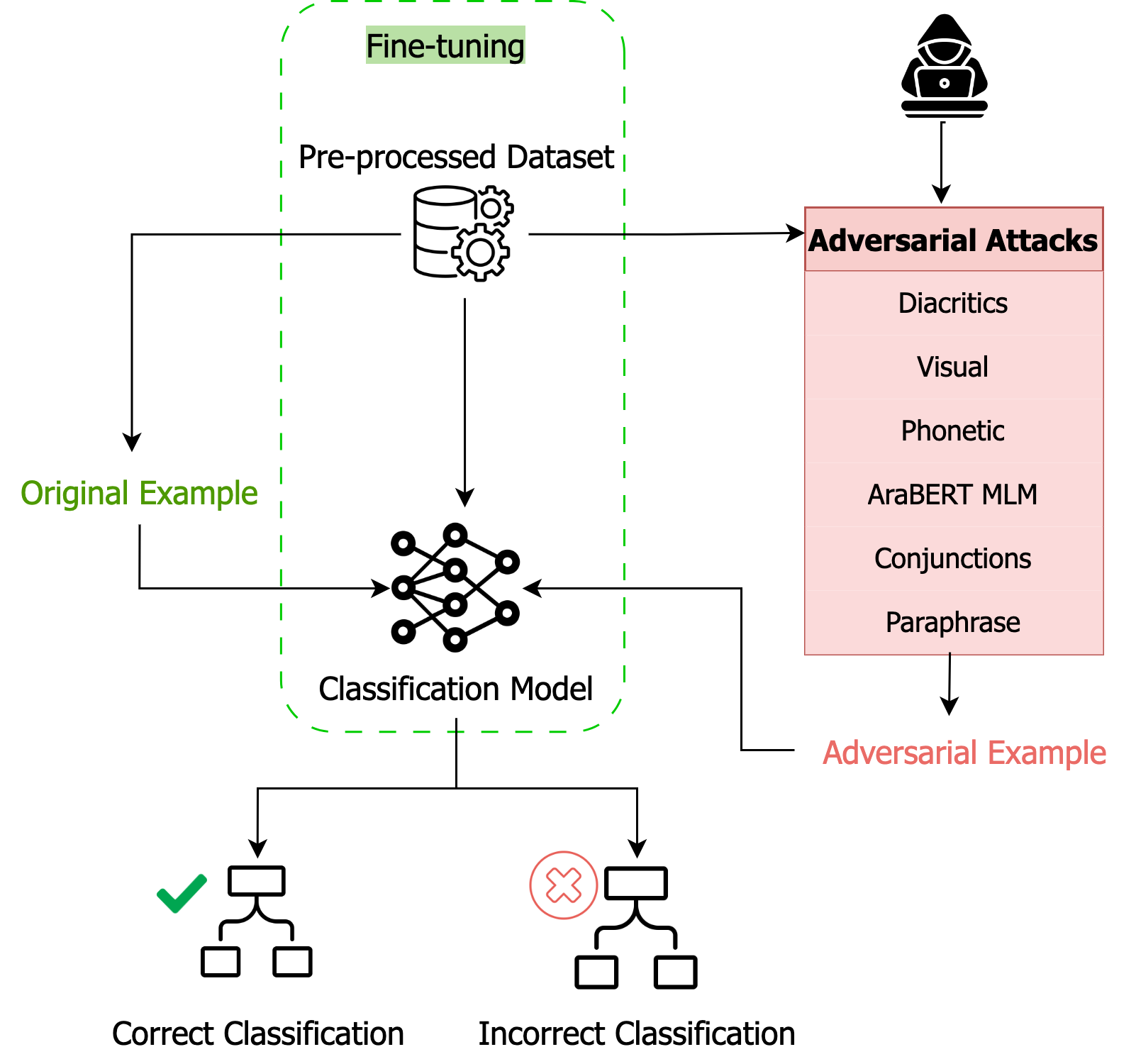}
    \caption{Attacks Framework.}
    \label{framework}
\end{figure}

Our approach evaluates the adversarial robustness of victim models against non-targeted black-box attacks. Adversarial robustness refers to a model’s ability to maintain performance when exposed to perturbed (adversarial) input samples. In the black-box setting, the attacker has no access to the internal architecture or parameters of the victim model and can only observe its outputs. The goal behind the attacks is to generate adversarial examples from the constructed dataset in a way that will lead to an incorrect output by the unfortified victim models. Having an original example $x$, where the model correctly classifies the input $f(x)=y$, each attack's algorithm generates a perturbed example $x'$ aimed to lead the victim models to missclassification such that $f(x')\neq y$. The framework of the attacks is demonstrated in Figure \ref{framework}. The models are trained on the pre-processed dataset to perform binary classification. Then, a subset of the dataset is manipulated through six attacks strategies to generate adversarial samples. The perturbed samples are tested on each model to evaluate the adversarial robustness.

\subsection{Attacks}
Six types of customized black-box and non-targeted Arabic attacks from \cite{alajmi2024evaluating} will be tested on each of the victim models. Three of the attacks are on a character-level (edit-based), two of them are on a word-level (importance-based), and one is on a sentence-level (paraphrase-based). The edit-based attacks do not take regard any character importance information in the perturbation process, whereas importance-based calculates the importance of every word to the prediction of the model and perturbs in a descending order of importance score. This is done through the use of SHapley Additive exPlanations (SHAP) that is based on game theory to interpret and score the words within the input sequence that lead to a model's decision \cite{shap}. To generate word‐level adversarial attacks, SHAP is used to compute the importance scores of individual words (see examples in Appendix~\ref{secA1}). Specifically, the \texttt{shap.Explainer} class was used with a token masker and \texttt{fixed\_context=1}, which masks one token at a time while preserving the rest of the sentence. This implicitly defines the baseline input as the unmasked context, eliminating the need for an explicit reference input. Once SHAP values are obtained for each token, they are sorted by magnitude and the most influential token is selected as the perturbation target, following the methodology of \cite{alajmi2025evaluating}. This procedure yields targeted, minimally invasive adversarial examples.

\begin{figure}[h!]
    \centering
    \includegraphics[width=0.8\linewidth]{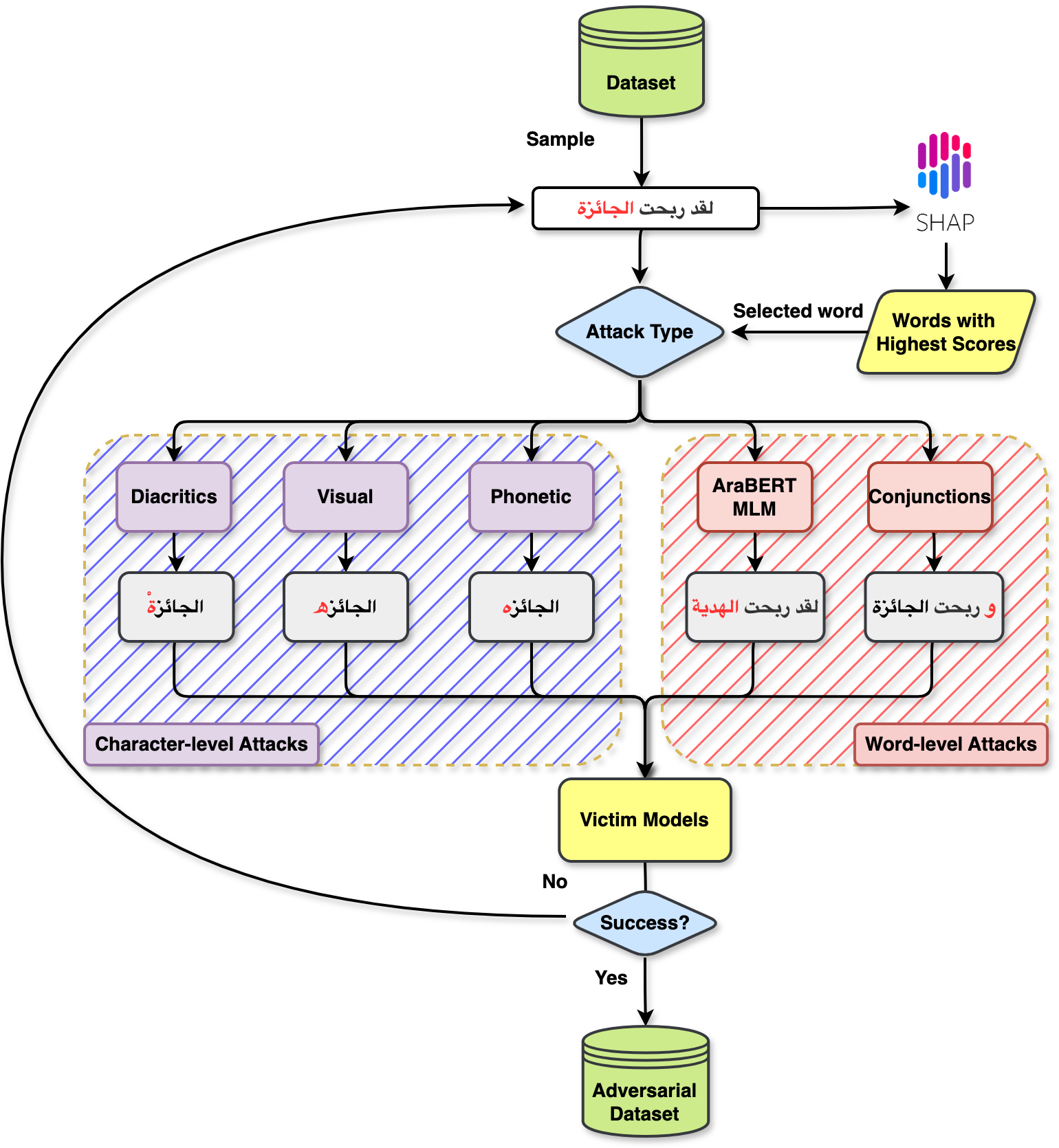}
    \caption{Flowchart Demonstrating The Character-level and Word-level Attacks Strategies.}
    \label{rob}
\end{figure}

Figure \ref{rob} demonstrates the attacks techniques using an illustrative example, while Table \ref{Attacks Examples} provides an example of each attack method. The details of constructing the adversarial samples are as follows:
\begin{itemize}

\item \textbf{Diacritics}: Arabic diacritics are vowels that can shift the meaning of words, where the insertion of different diacritics can cause one word to multiple meanings. For instance, \raisebox{-0.9ex}{\includegraphics[height=2.8ex]{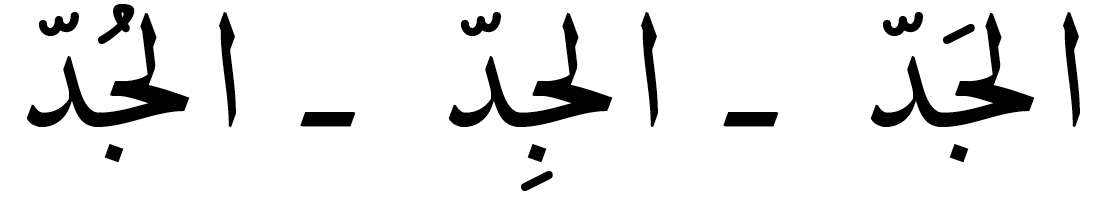}} are words with the same letters but denote three entirely different meanings in Arabic, highlighting the complexity of the language. This attack incorporates the insertion of a variety of Arabic diacritics to deceive the victim models.

\item \textbf{Visual}: The Arabic language consists of distinct characters that are visually similar such as \raisebox{-0.9ex}{\includegraphics[height=2.8ex]{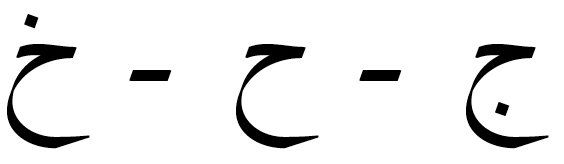}} which are only differentiated by one dot. Hence, the attacks substitutes the input characters with visually similar ones.

\item \textbf{Phonetic}: This attack targets the dialectal differences across Arabic-speaking regions. With roughly 25 dialects in the Arab world, a word with the same semantic meaning can be written differently in various countries, particularly when it comes to male and female pronouns. The strategy involves performing character replacements to change the dialect of the original sequence.

\item \textbf{AraBERT MLM}: Performs word substitutions with synonyms generated by AraBERT Masked Language Model (MLM), where the substitution begins with the word that has the highest SHAP score.

\item \textbf{Conjunctions}: Replacement of Arabic conjunctions with semantically similar ones starting with the ones that have higher SHAP scores.

\item \textbf{Paraphrase}: Rephrasing Arabic sentences similar to \cite{alajmi2025evaluating}, while preserving the semantics of the original sequence using the abstractive summarization AraT5 model in \cite{bani2023arabic}. 
\end{itemize}
\begin{sidewaystable}
\begin{center}
\begin{minipage}{\textheight}
    \begin{tabular*}{\textwidth}{@{\extracolsep\fill}lccl} 
 \toprule
 \textbf{Attack} & \textbf{Attack Level} & \textbf{Perturbation Strategy} & \textbf{Example}    \\
 \midrule
No Attack & - & - &
  \raisebox{-0.9ex}{\includegraphics[height=2.8ex]{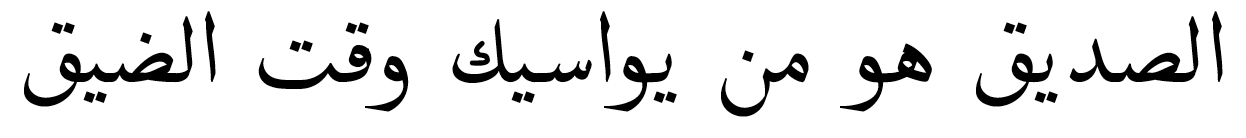}}\\
Diacritic & Character-level & Edit-based &
  \raisebox{-0.9ex}{\includegraphics[height=2.8ex]{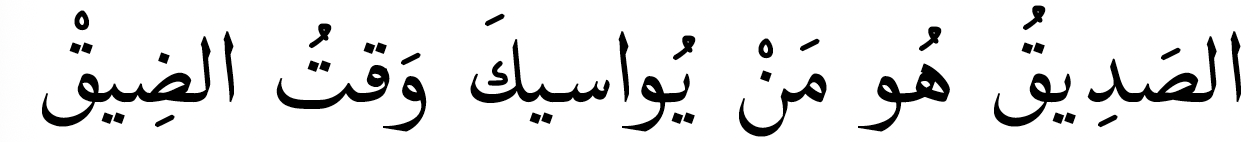}}\\
Visual & Character-level & Edit-based &
  \raisebox{-0.9ex}{\includegraphics[height=2.4ex]{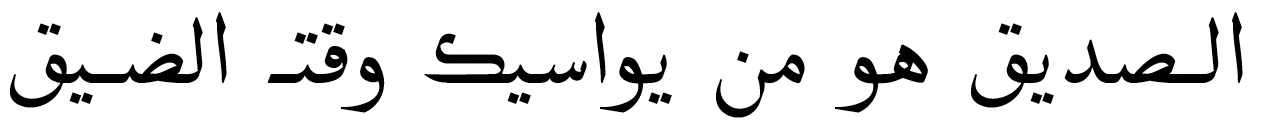}}\\
Phonetic & Character-level & Edit-based &
  \raisebox{-0.9ex}{\includegraphics[height=2.8ex]{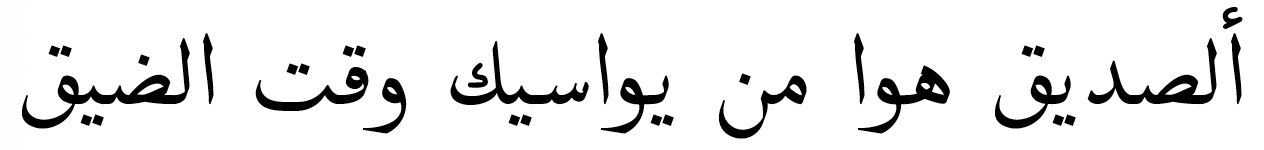}}\\
AraBERT MLM & Word-level & Importance-based &
  \raisebox{-0.9ex}{\includegraphics[height=2.4ex]{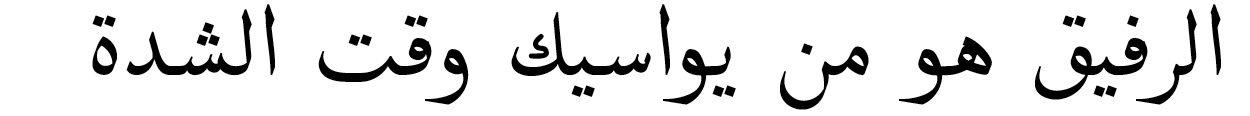}}\\
Conjunctions & Word-level & Importance-based &
  \raisebox{-0.9ex}{\includegraphics[height=2.8ex]{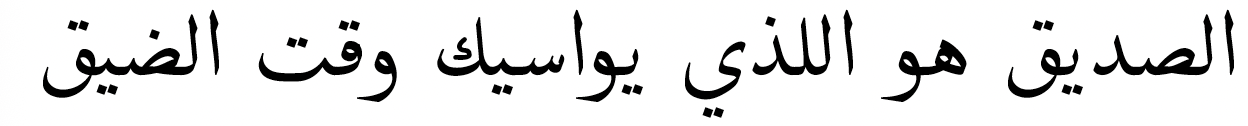}}\\
Paraphrase & Sentence-level & Edit-based &
  \raisebox{-0.9ex}{\includegraphics[height=2.4ex]{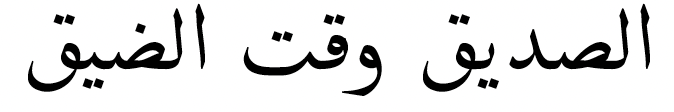}}\\
\bottomrule
\end{tabular*}
\caption{Examples of the attacks.}
    \label{Attacks Examples}
    \end{minipage}
    \end{center}
\end{sidewaystable}

\section{Experimental Results and Analysis}\label{sec4}

After data pre-processing and class balancing techniques, the victim models were trained on the constructed dataset for binary sentiment analysis through a unified training approach and hyperparameters selection for objective evaluation. The pre-trained models are loaded from \texttt{huggingface} and finetuned with \texttt{early stopping} to prevent them from overfitting. In order to properly evaluate the adversarial robustness of the unfortified models the following metrics will be incorporated:

\begin{itemize}

\item  \textit{Accuracy}: The number of correct predictions over the model's total predictions as shown in Equation \ref{accuracy}, where TP denotes the number of true positives, and TN denotes the number of true negatives, and FP represent the number of false positives, whereas FN denotes the number of false negatives.
\begin{equation}
    Accuracy = \frac{TP + TN}{TP + TN + FP + FN}
        \label{accuracy}
        \end{equation}

\item \textit{Attack Success Rate (ASR)}: A metric that is commonly used to evaluate the effectiveness of adversarial attacks. As represented in Equation \ref{equationasr}, ASR measures the percentage of successful adversarial samples ($N_{\text{successful}}$) over the total number of adversarial samples ($N_{\text{total}}$). In our experiments, ($N_{\text{successful}}$) denotes the number of adversarial examples for which the model’s prediction on the perturbed input differs from its original correct prediction on the clean input. Samples where the clean-input prediction was already incorrect are excluded from ASR computation.
\begin{equation}
    \text{ASR} = \left( \frac{N_{\text{successful}}}{N_{\text{total}}} \right) \times 100\%
    \label{equationasr}
\end{equation}
\item  \textit{Perturbation Rate}: To measure the perturbation distance between the original and perturbed examples, Levenshtein distance is selected due to its efficiency in including all perturbation operations on sequences of different sizes. It is shown in Equation \ref{levin} where the Levenshtein distance is $d_{i,j}$ between the original example $A$ with the first character $x_i$ and the perturbed example $B$ with the first character $y_j$. The function $I$ returns 1 if $x_i \neq y_j$ \cite{levenshtein1966binary}. 

\begin{equation}
    d_{x,y}(i,j) =
    \begin{cases}
        i & \text{if } j = 0 \\
        j & \text{if } i = 0 \\
        \min \left\{
            \begin{aligned}
                &d_{i-1,j} + 1, \\
                &d_{i,j-1} + 1, \\
                &d_{i-1,j-1} + I(x_i \neq y_j)
            \end{aligned}
        \right.
    \end{cases}
    \label{levin}
\end{equation}
 \item  \textit{Cosine Similarity}: The cosine angle between the original sequence vector A and perturbed sequence vector B (Equation \ref{equationcosine}). The cosine value increases with higher similarities between the original and adversarial examples.
\begin{equation}
\cos(\theta) = \frac{\mathbf{A} \cdot \mathbf{B}}{||\mathbf{A}|| ||\mathbf{B}||}
\label{equationcosine}
\end{equation}
 \item  \textit{Universal Sentence Encoder (USE)}: A more recent approach that extends Cosine Similarity by using arccos to find the angular distance between the original sequence vector A and perturbed sequence vector B \cite{cer2018universal} as shown in Equation \ref{equationuse}. High USE score indicates similar original and adversarial samples, which means that the semantic similarity is preserved.

\begin{equation}
\text{sim}(\mathbf{A}, \mathbf{B}) = 1 - \frac{\arccos\left(\cos(\theta)\right)}{\pi}
\label{equationuse}
\end{equation}

        \end{itemize}

\subsection{Adversarial Attacks Results}

\begin{figure}[h]
\begin{subfigure}{0.49\textwidth}
    \includegraphics[width=\linewidth]{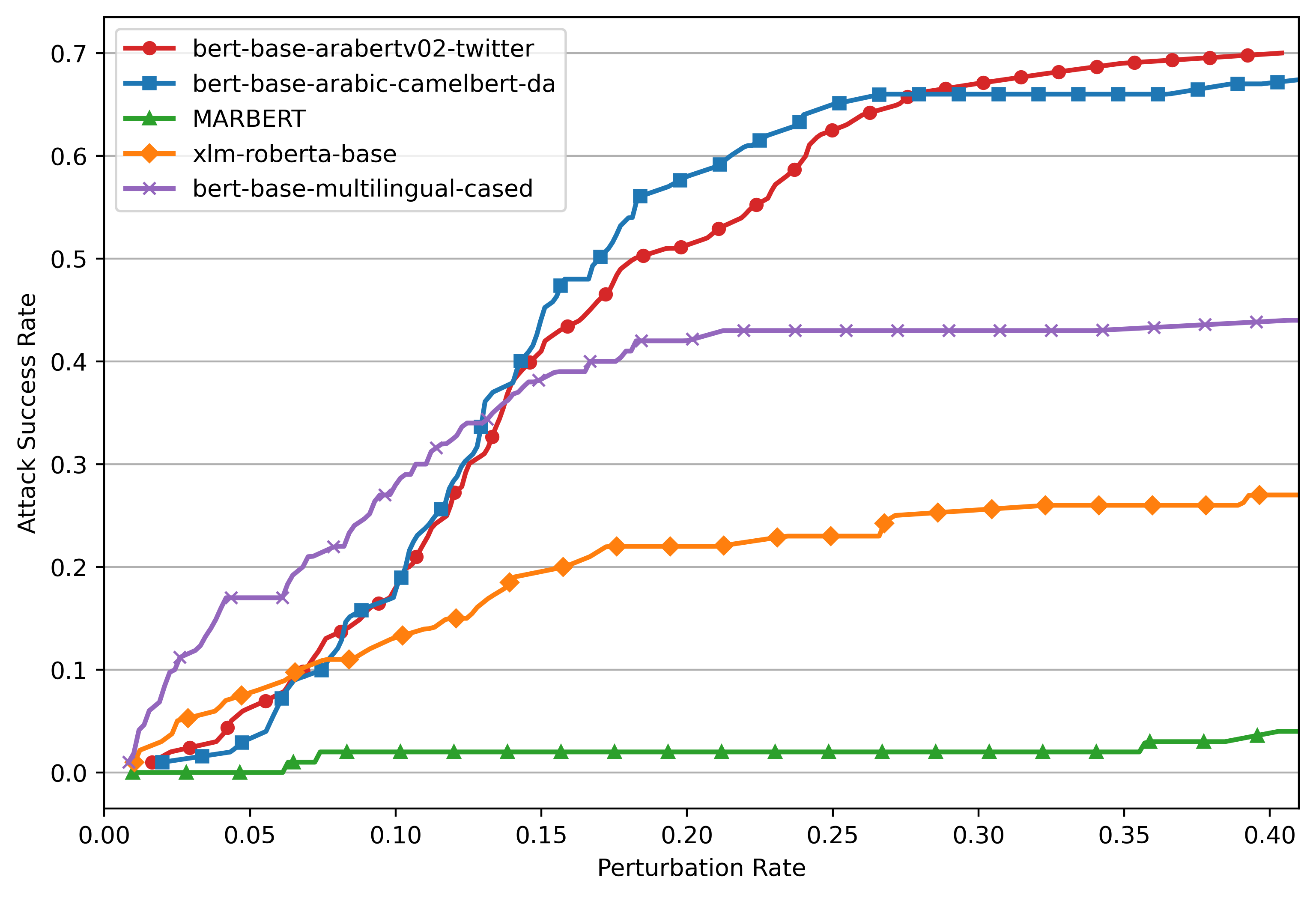} 
\end{subfigure}
\hfill 
\begin{subfigure}{0.49\textwidth}
    \includegraphics[width=\linewidth]{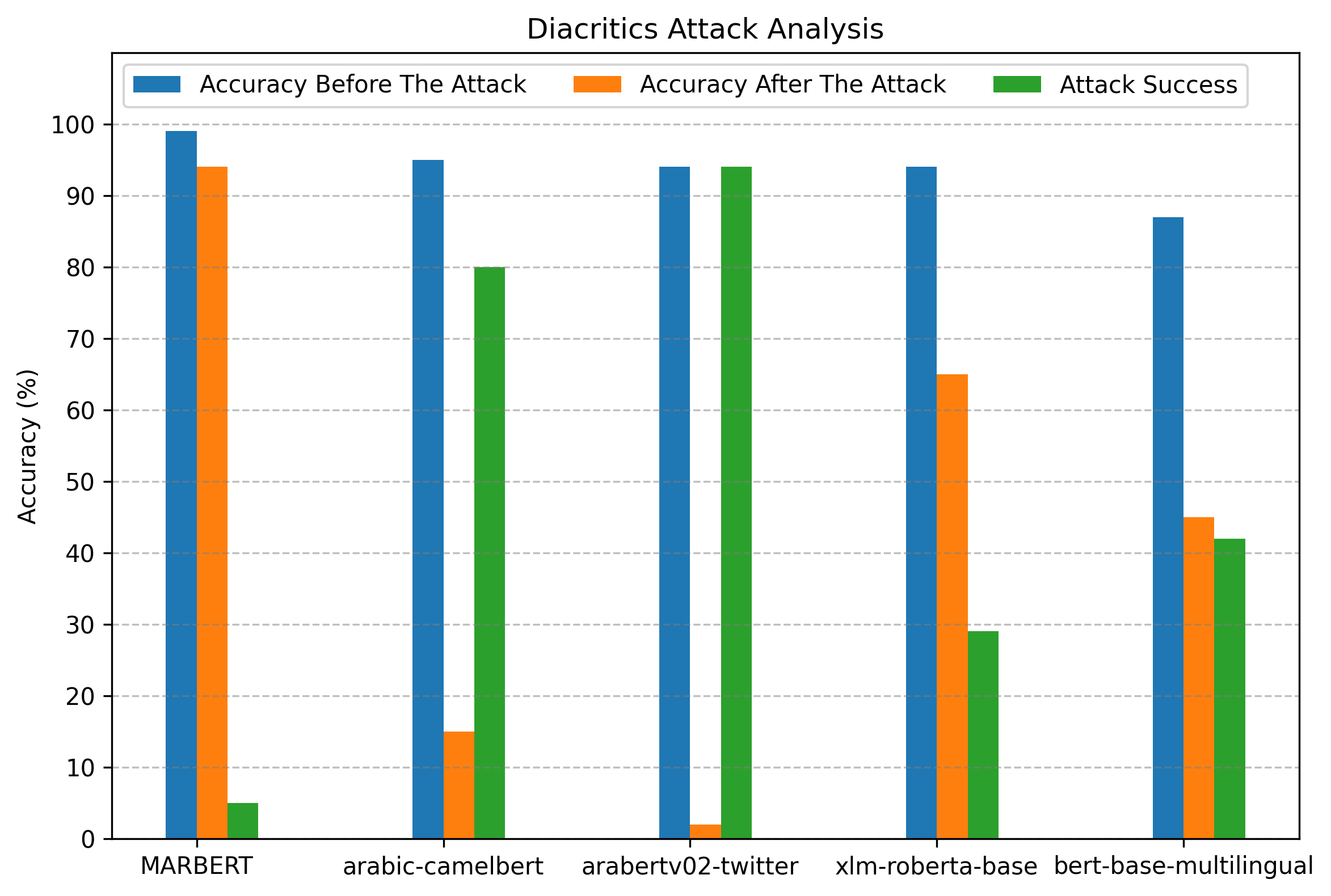}
\end{subfigure}
\caption{The results of diacritics manipulation on the models.}
\label{fig:Diacritics}
\end{figure}

The accuracies of the models on the test set are: 99\% for MARBERT, 95\% for CaMeLBERT, 94\% for AraBERT and XLM-T, and 87\% for mBERT. Table \ref{Attacks Results Diacritics} demonstrates the effects of diacritics insertions on the performance of the victim models. It is shown that MARBERT is the most robust against this type of attacks with 94\% post-attack accuracy. On the other hand, AraBERT is the most vulnerable to diacritics manipulation with an accuracy decrease of 92\% followed by CaMeLBERT (80\% decrease). 
When it comes to perturbation rate, mBERT model adversarial samples modify an average of only 10\% of the original samples, while retaining the highest cosine and USE similarity scores of 90\% and 68\%, respectively. The attack success rates plotted against the perturbation rates of the victim models are shown in Figure \ref{fig:Diacritics} as well as the pre-attack and post-attack accuracies. The experimental results of diacritics manipulation proves that character-level insertions attacks are feasible. This kind of attacks is not affected by
the limitations of character or word perturbations, especially that it is hard to be detected by passive defense approaches like grammar checkers, while preserving the semantics on the input \cite{formento-etal-2023-using}.

\begin{sidewaystable}
\begin{center}
\begin{minipage}{\textheight}
\begin{tabular*}{\textwidth}{@{\extracolsep\fill}p{4cm}p{2cm}p{2cm}p{1.5cm}p{2cm}p{2cm}p{2cm}} 
 \toprule
      Model & Pre-attack Accuracy & Post-attack Accuracy & Decrease & Perturbation Rate & Cosine \newline Similarity & USE \newline Similarity \\
          \midrule
AraBERT & 94\% &  2\% & \textbf{92\%} & 15\% & 84\% & 56\% \\
MARBERT & \textbf{99\%} & \textbf{94\% }& 5\% & 28.5\% & 57\% & 55\% \\
CaMeLBERT & 95\% & 15\% & 80\% & 15.5 \%& 79\% & 55\% \\
mBERT &  87 \% & 45\% & 42\% & \textbf{10\%} & \textbf{90\%} & \textbf{68\%} \\  
XLM-T & 94\% & 65\% & 29\% & 33\% & 53\% & 52\% \\

\bottomrule
    \end{tabular*}
     \caption{Diacritics Attacks Evaluation Results}
    \label{Attacks Results Diacritics}
\end{minipage}
\end{center}
\end{sidewaystable}

\begin{figure}[h]
\begin{subfigure}{0.49\textwidth}
    \includegraphics[width=\linewidth]{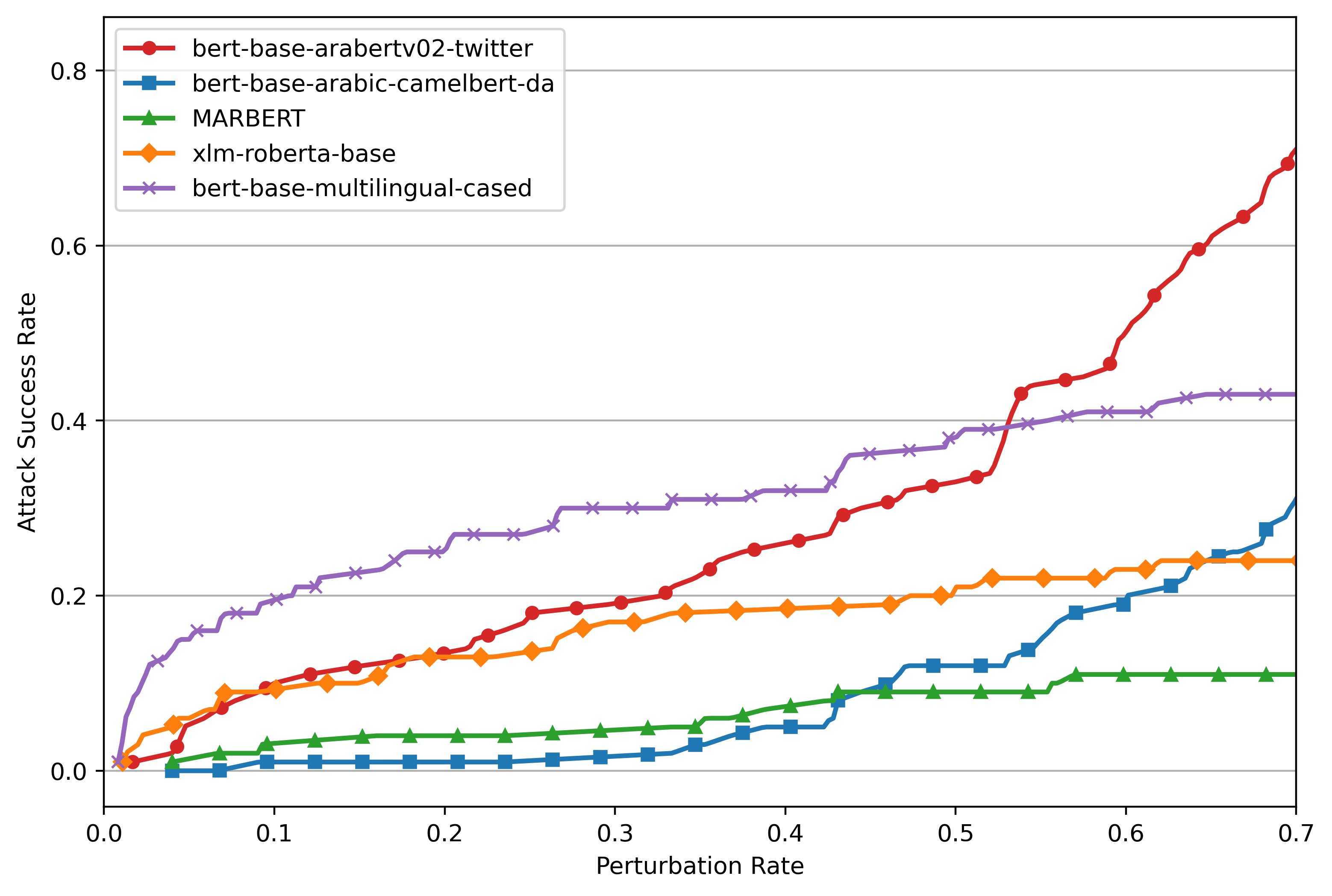} 
\end{subfigure}
\hfill 
\begin{subfigure}{0.49\textwidth}
    \includegraphics[width=\linewidth]{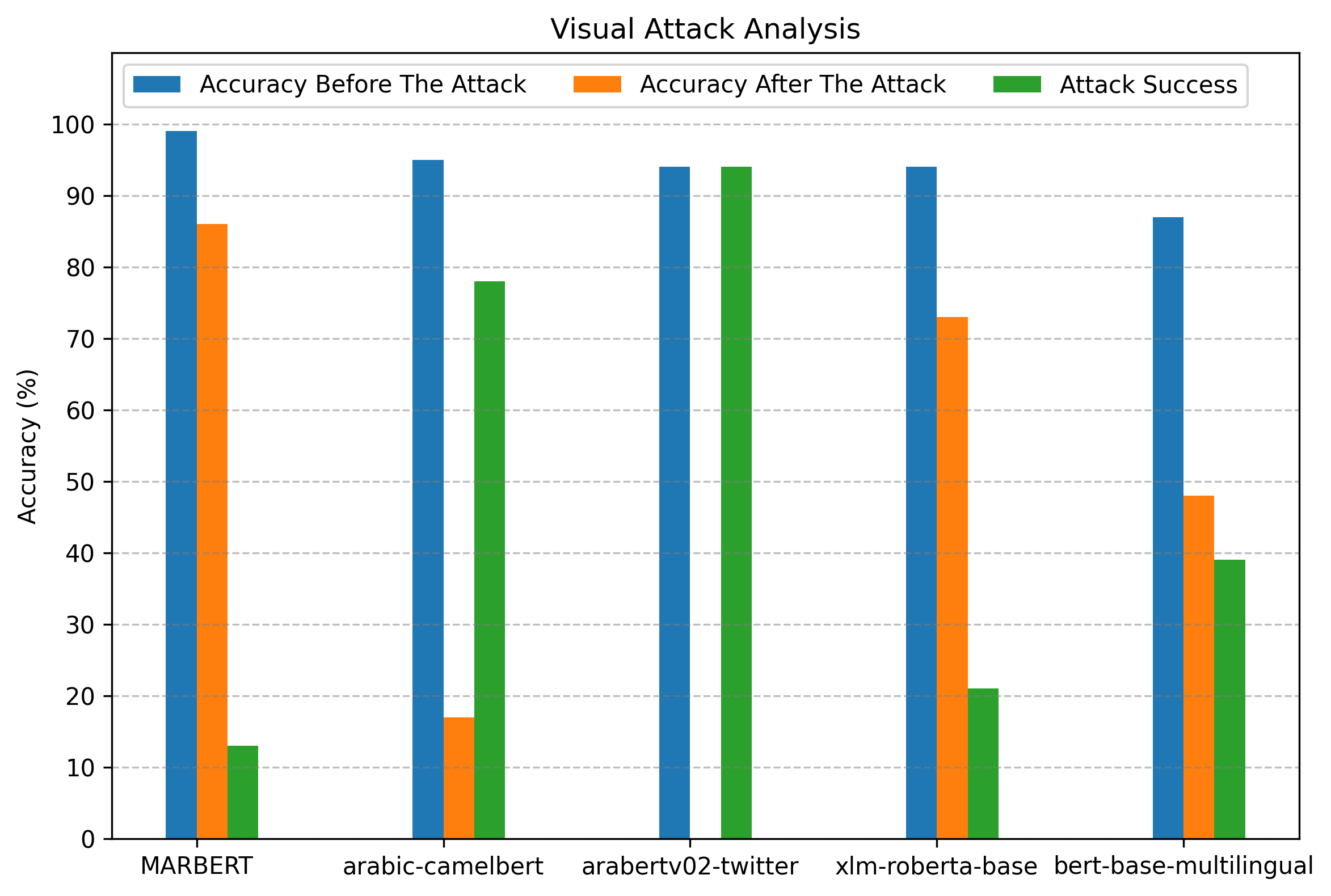}
\end{subfigure}
\caption{The victim models' performance evaluation after the visual attack.}
\label{fig:Visual}
\end{figure}

\begin{sidewaystable}
\begin{center}
\begin{minipage}{\textheight}
\begin{tabular*}{\textwidth}{@{\extracolsep\fill}p{4cm}p{2cm}p{2cm}p{1.5cm}p{2cm}p{2cm}p{2cm}} 
 \toprule
      Model & Pre-attack Accuracy & Post-attack Accuracy & Decrease & Perturbation Rate & Cosine \newline Similarity & USE \newline Similarity \\
          \midrule
AraBERT & 94\% & 0\% & \textbf{94\%} & 31\% & 26\% & 56\% \\
MARBERT  & \textbf{99\% }& \textbf{86\% }& 13\% & 45\% & 12\% & 50\% \\
CaMeLBERT & 95\% & 17\% & 78\% & 38\% & 31\% & 63\% \\
 mBERT & 87 \% & 48\% & 39\% & \textbf{18\% } & \textbf{53\%} & \textbf{72\%} \\  
XLM-T & 94\% & 73\% & 21\% & 43\%  & 27\% & 58\%\\

\bottomrule
    \end{tabular*}
      \caption{Visual Attacks Evaluation Results}
    \label{Attacks Results Visual}
\end{minipage}
\end{center}
\end{sidewaystable}

The results of applying visual attack on the victim models are indicated in Table \ref{Attacks Results Visual}. Again, AraBERT is the weakest to visual replacement with 0\% post-attack accuracy, while MARBERT is the most robust with 86\% post-attack accuracy as shown in Figure \ref{fig:Visual}. 
MARBERT may be less susceptible to visual replacement attacks due to its diverse and extensive pre-training data, which includes a wide range of Arabic dialects. The pre-training corpus also included tweets that contained at least three Arabic words but did not exclude non-Arabic content. As a result, the model was exposed to a variety of mixed-script text. This exposure to noisy and heterogeneous data during pre-training likely enhanced MARBERT’s ability to handle such perturbations, contributing to its robustness. Moreover, this kind of attack results in high perturbation rate for all the models due to the character swapping operation which is more expensive in terms of the adversary budget (character deletion and insertion) than the diacritics insertion attack. However, it only takes a perturbation rate of 18\% for mBERT to drop its accuracy by 39\%, while maintaining the highest cosine and USE similarity scores. 

\begin{sidewaystable}
\begin{center}
\begin{minipage}{\textheight}
\begin{tabular*}{\textwidth}{@{\extracolsep\fill}p{4cm}p{2cm}p{2cm}p{1.5cm}p{2cm}p{2cm}p{2cm}} 
 \toprule
      Model & Pre-attack Accuracy & Post-attack Accuracy & Decrease & Perturbation Rate & Cosine \newline Similarity & USE \newline Similarity \\
\midrule
AraBERT & 94\% & 91\% & 3\% & 28\% & 33\% & 62\%  \\
MARBERT & \textbf{99\% }& \textbf{97\% }& 2\% & 37\% & 33\% & 63\%\\
CaMeLBERT & 95\% & 64\% & \textbf{31\%} & 33\% & 30\% & 61\%\\
mBERT & 87 \% & 65\% & 22\% & \textbf{17\%} & \textbf{50\%} & \textbf{70\%} \\  
XLM-T & 94\% & 80\% & 14\% & 36\% & 34\% & 64\% \\

\bottomrule
    \end{tabular*}
      \caption{Phonetic Attacks Evaluation Results}
    \label{Attacks Results Phonetic}
\end{minipage}
\end{center}
\end{sidewaystable}

\begin{figure}[h]
\begin{subfigure}{0.49\textwidth}
    \includegraphics[width=\linewidth]{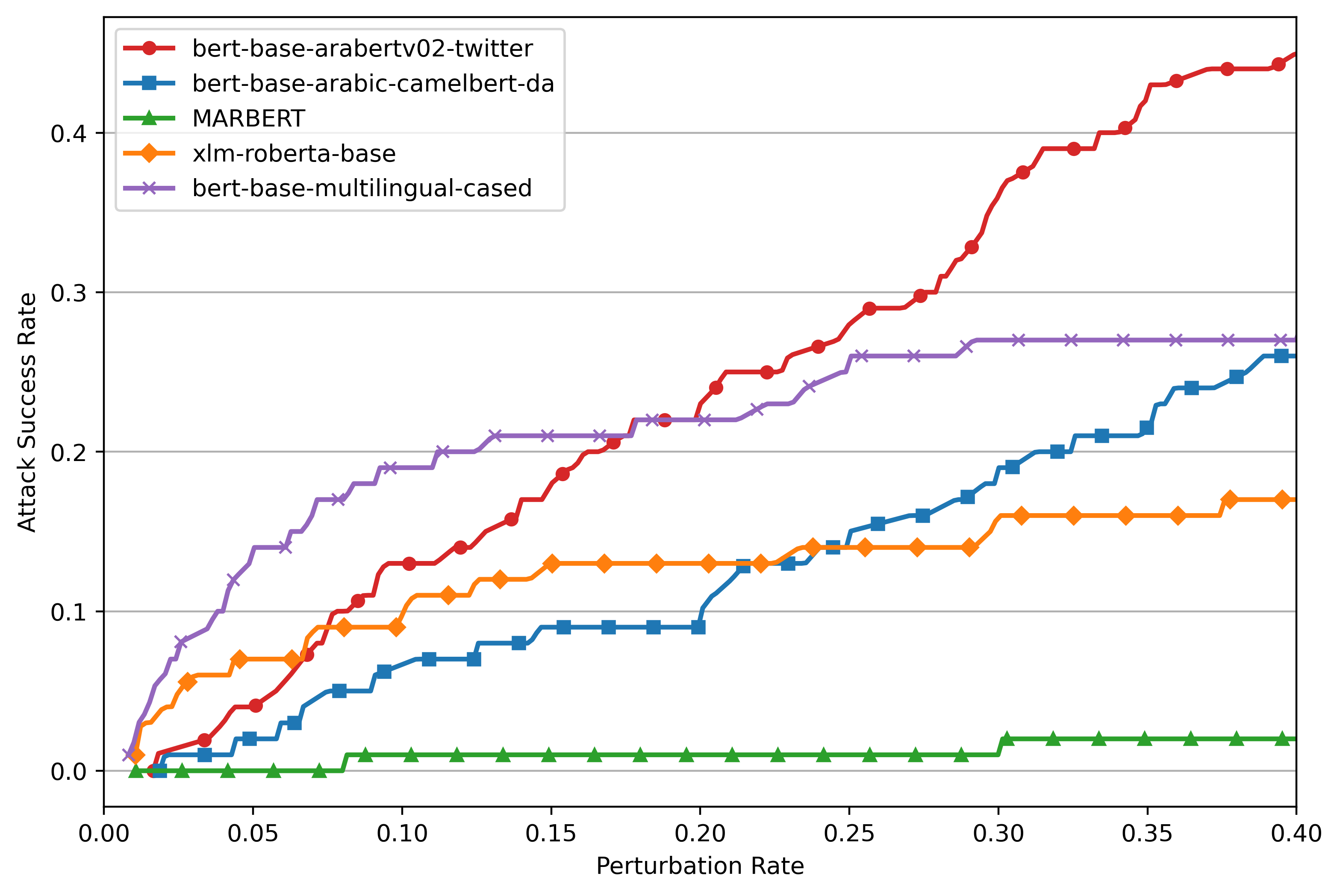} 
\end{subfigure}
\hfill 
\begin{subfigure}{0.49\textwidth}
    \includegraphics[width=\linewidth]{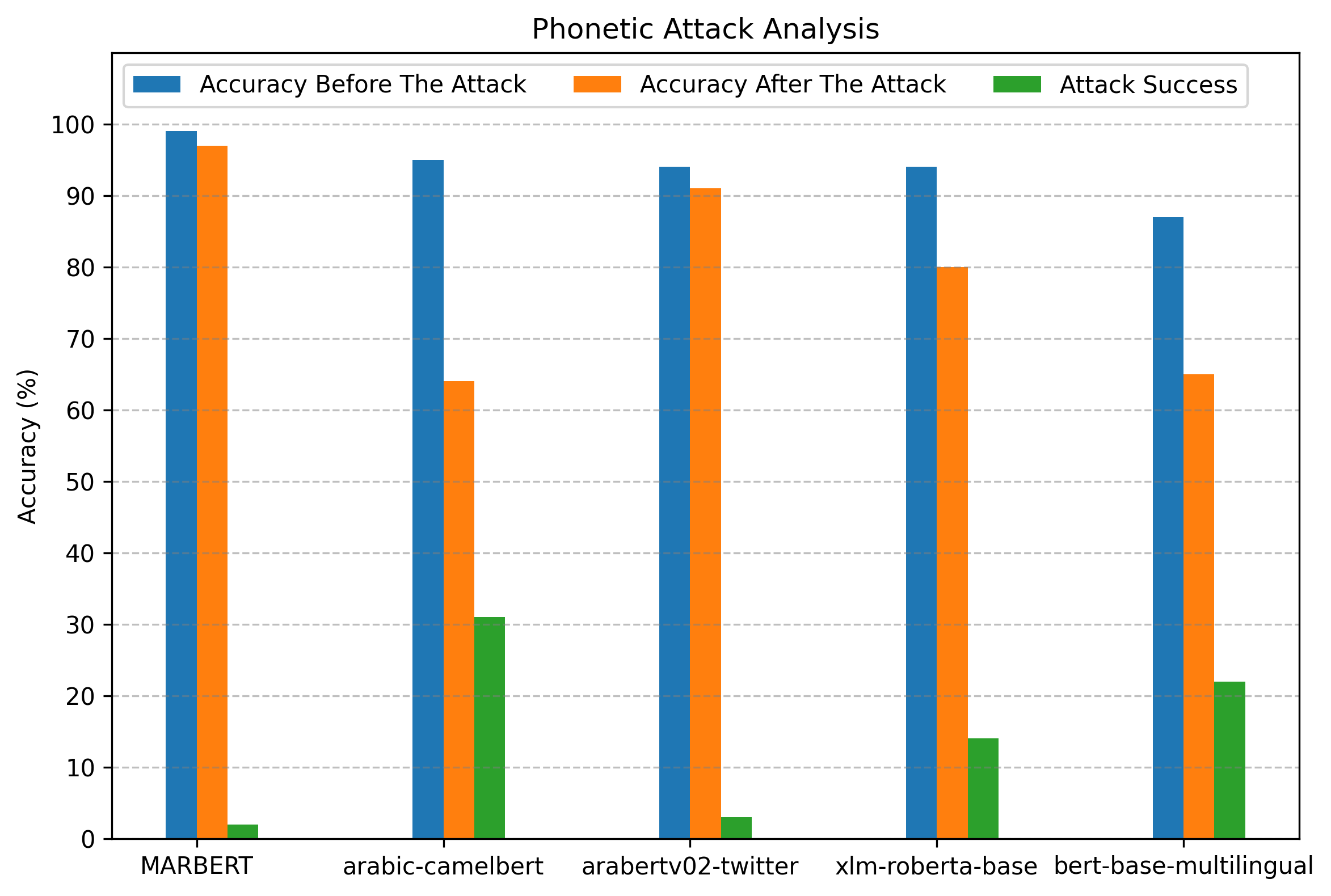}
\end{subfigure}
 \caption{The effects of phonetic character substitution on the victim models.}
\label{fig:Phonetic}
\end{figure}

As for the last character-level attack, Table \ref{Attacks Results Phonetic} and Figure \ref{fig:Phonetic} surprisingly show that CaMeLBERT's performance proves to be the lowest among the victim models after performing phonetic character substitutions (31\% drop in accuracy). Although this model is pre-trained on a datasets that include a variety of dialects, this finding reveals a generalization problem when it comes to unseen Arabic data. On the other hand, MARBERT, shows good generalization capability as it is barely affected with the phonetic attack with only 2\% decrease in its pre-attack accuracy. In addition, the second most vulnerable model to dialect manipulations is mBERT, in which the examples sustain the highest cosine and USE similarity results, while having the lowest perturbation rate of 17\%. This is relevant to mBERT's pre-training data which consists only of MSA with no dialectal Arabic inclusion.

\begin{sidewaystable}
\begin{center}
\begin{minipage}{\textheight}
\begin{tabular*}{\textwidth}{@{\extracolsep\fill}p{4cm}p{2cm}p{2cm}p{1.5cm}p{2cm}p{2cm}p{2cm}} 
 \toprule
      Model & Pre-attack Accuracy & Post-attack Accuracy & Decrease & Perturbation Rate & Cosine \newline Similarity & USE \newline Similarity \\
\midrule
AraBERT & 94\% & \textbf{81\%} & 13\% & 70\% & 38\% & 61\% \\
MARBERT & \textbf{99\%} & 60\% & 39\% & 81\% & 27\% & 59\% \\
CaMeLBERT & 95\% & 70.7\% & 24.3\% & \textbf{64\% }& 29\% &  \textbf{62\%} \\
mBERT& 87 \% & 47.8\% & 39.2\% & 69\% & \textbf{36\%} & 61\% \\  
XLM-T& 94\% & 50\% & \textbf{44\%} & 82\% & 25\% & 58\% \\

\bottomrule
    \end{tabular*}
  \caption{MLM Attacks Evaluation Results}
    \label{Attacks Results MLM}
\end{minipage}
\end{center}
\end{sidewaystable}

With the use of SHAP and AraBERT MLM to generate replacements for the most significant words within the original sample, the impact on the target models can be seen in Table \ref{Attacks Results MLM}. The attack was successful the most on XLM-T, which exhibited an overall accuracy decrease of 44\% with the highest perturbation rate, followed by mBERT (highest Cosine Similarity) and MARBERT, which both approximately have the same amount of drop in accuracy by 39\%. Meanwhile, AraBERT had the highest post-attack accuracy of 81\% revealing that it is less vulnerable to similar word substitution than the other tested models.

\begin{sidewaystable}
\begin{center}
\begin{minipage}{\textheight}
\begin{tabular*}{\textwidth}{@{\extracolsep\fill}p{4cm}p{2cm}p{2cm}p{1.5cm}p{2cm}p{2cm}p{2cm}} 
 \toprule
      Model & Pre-attack Accuracy & Post-attack Accuracy & Decrease & Perturbation Rate & Cosine \newline Similarity & USE \newline Similarity \\
\midrule
AraBERT & 94\% & 36\% & \textbf{58\%} & 29\% & 96\% & 51\% \\
MARBERT &\textbf{ 99\%} & \textbf{97.5\%} & 1.5\% & 25\% & 93\% & 51\%\\
CaMeLBERT  & 95\% & 73.5\% & 21.5\% & \textbf{24\%} & \textbf{96\%} & \textbf{52\%}\\
mBERT& 87 \% & 83\% & 4\% & 26\% & 94\% & 50\% \\  
XLM-T  & 94\% & 55\% & 39\% & 53\% & 94\% & 50\% \\
\bottomrule
    \end{tabular*}
 \caption{Conjunctions Attacks Evaluation Results}
    \label{Attacks Results Conjunction}
\end{minipage}
\end{center}
\end{sidewaystable}

\begin{figure}[h!]
\begin{subfigure}{0.49\textwidth}
    \includegraphics[width=\linewidth]{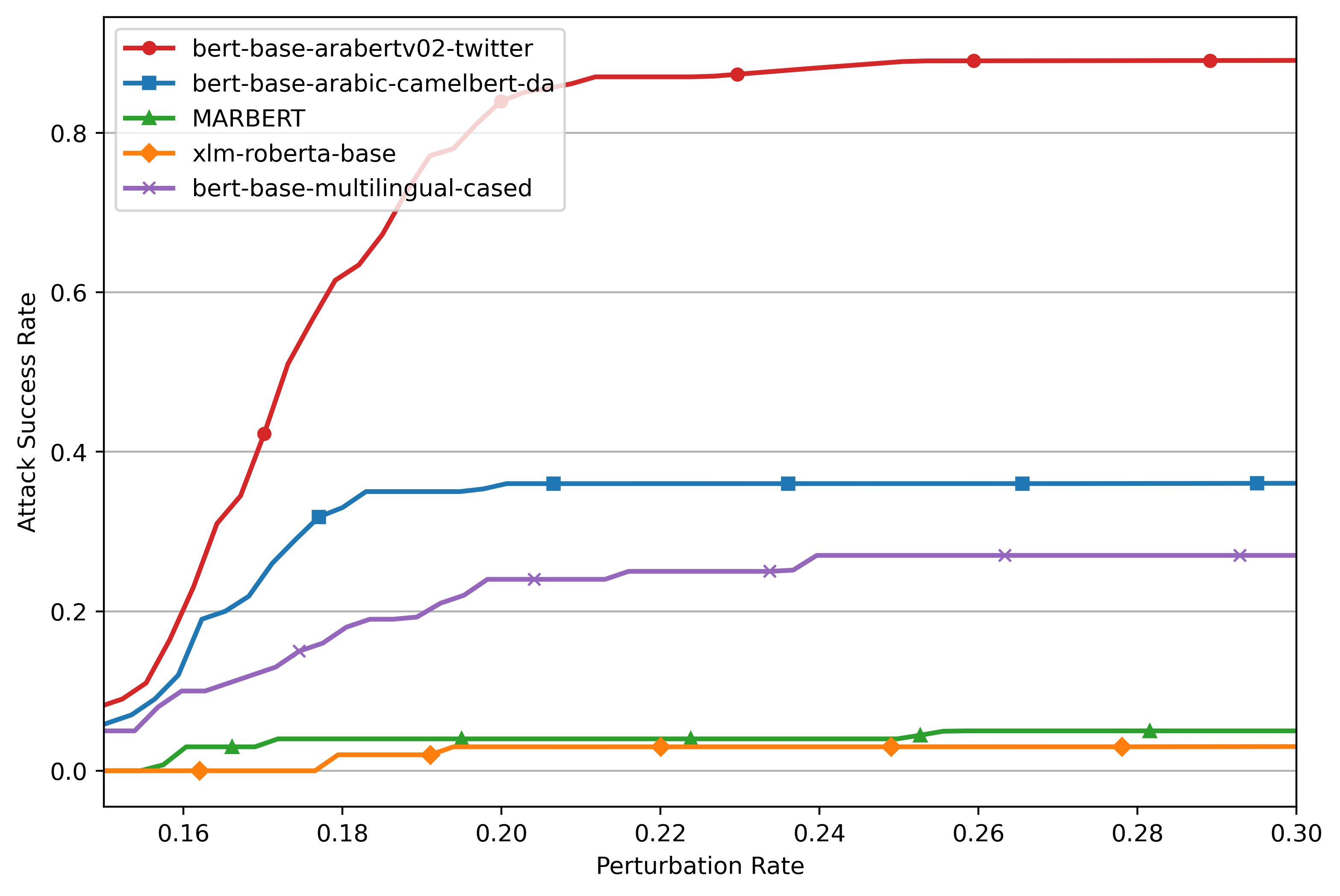} 
\end{subfigure}
\hfill 
\begin{subfigure}{0.49\textwidth}
    \includegraphics[width=\linewidth]{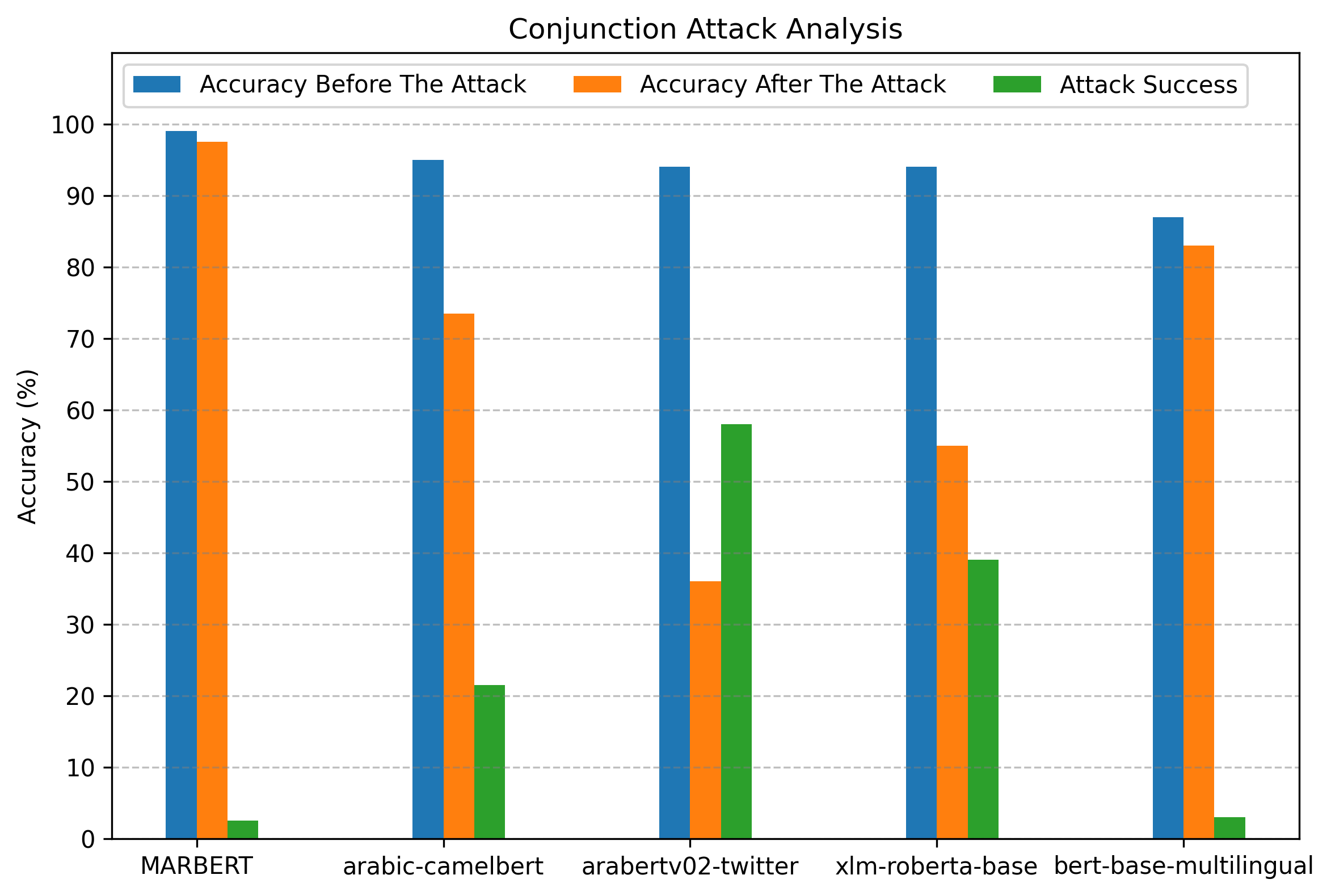}
\end{subfigure}
 \caption{The results obtained from conjunctions manipulation on the victim models.}
    \label{fig:Conj}
\end{figure}

Applying conjunctions manipulation on the victim models can notably impact their performance as seen in Figure \ref{fig:Conj}. It is observed in Table \ref{Attacks Results Conjunction} that AraBERT and XLM-T have endured the heaviest accuracy reduction by 58\% and 39\%, respectively. This can be related to the models reliance on understanding the syntactic structure and relationships conveyed by these conjunctions, especially that this attack technique involves altering conjunctions within the original example leading to confuse the models' understanding of different parts of the sentence. However, this time CaMeLBERT has the best scores when it comes to perturbation rate and similarity scores, showing that it is capable of understanding the text's overall meaning, even when the Arabic conjunctions are manipulated.

\begin{table}[h!]
    \centering
     \begin{tabular*}{0.9\textwidth}{@{\extracolsep\fill}lccc}
     \toprule
     Model & Pre-attack Accuracy & Post-attack Accuracy & Decrease \\
          \midrule
          
AraBERT & 94\% & 14\% & 80\%  \\
MARBERT & \textbf{99\%} & 14\% & \textbf{85\%} \\
CaMeLBERT & 95\% & 22\% & 73\% \\
mBERT& 87 \% & \textbf{28\%} & 59\% \\  
XLM-T  & 94\% & 11\% & 83\% \\
\bottomrule
    \end{tabular*}
    \caption{Paraphrase Attacks Evaluation Results}
    \label{Attacks Results Paraphrase}
    \end{table}

\begin{figure}[h!]
\begin{subfigure}{0.49\textwidth}
    \includegraphics[width=\linewidth]{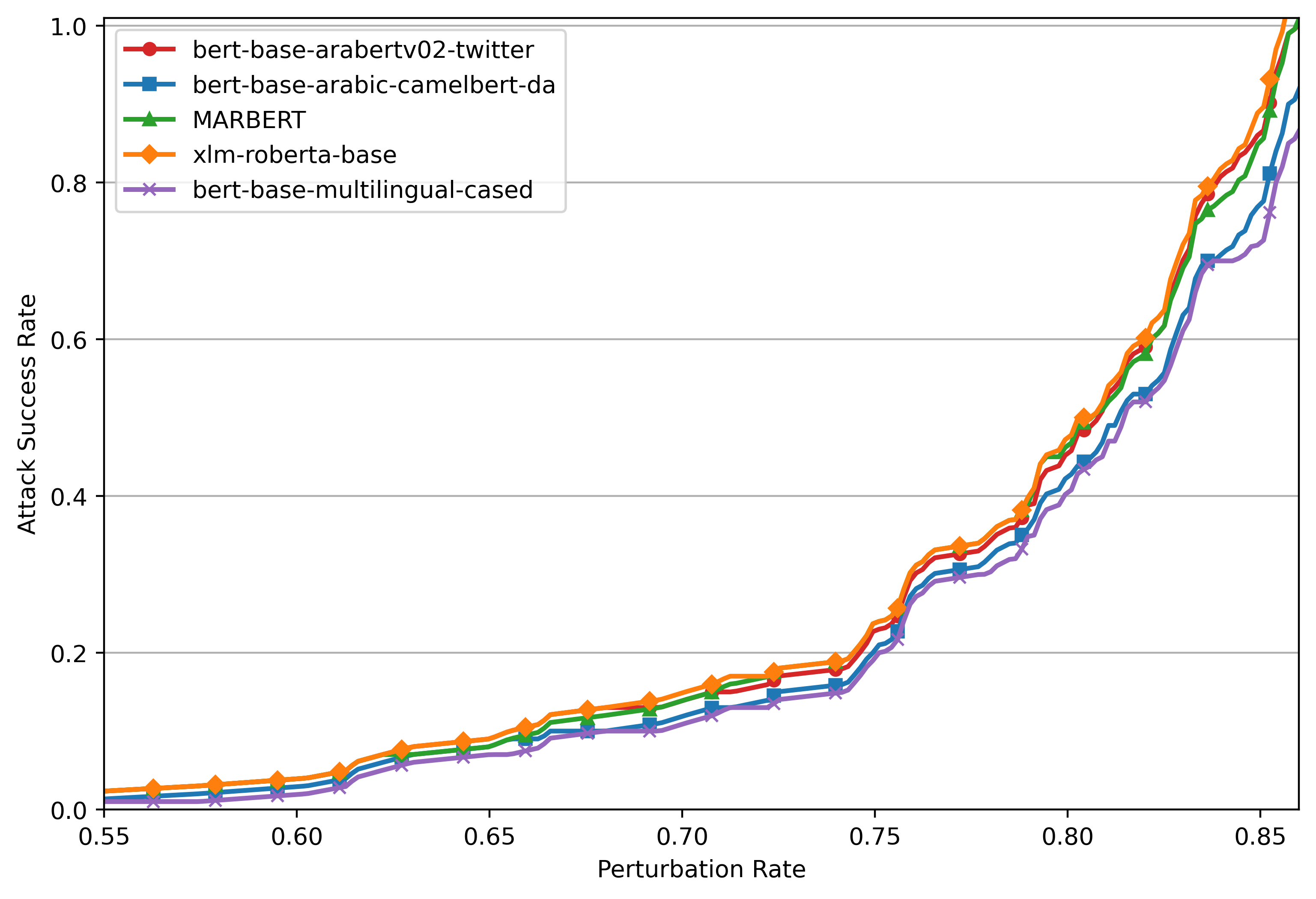} 
\end{subfigure}
\hfill 
\begin{subfigure}{0.49\textwidth}
    \includegraphics[width=\linewidth]{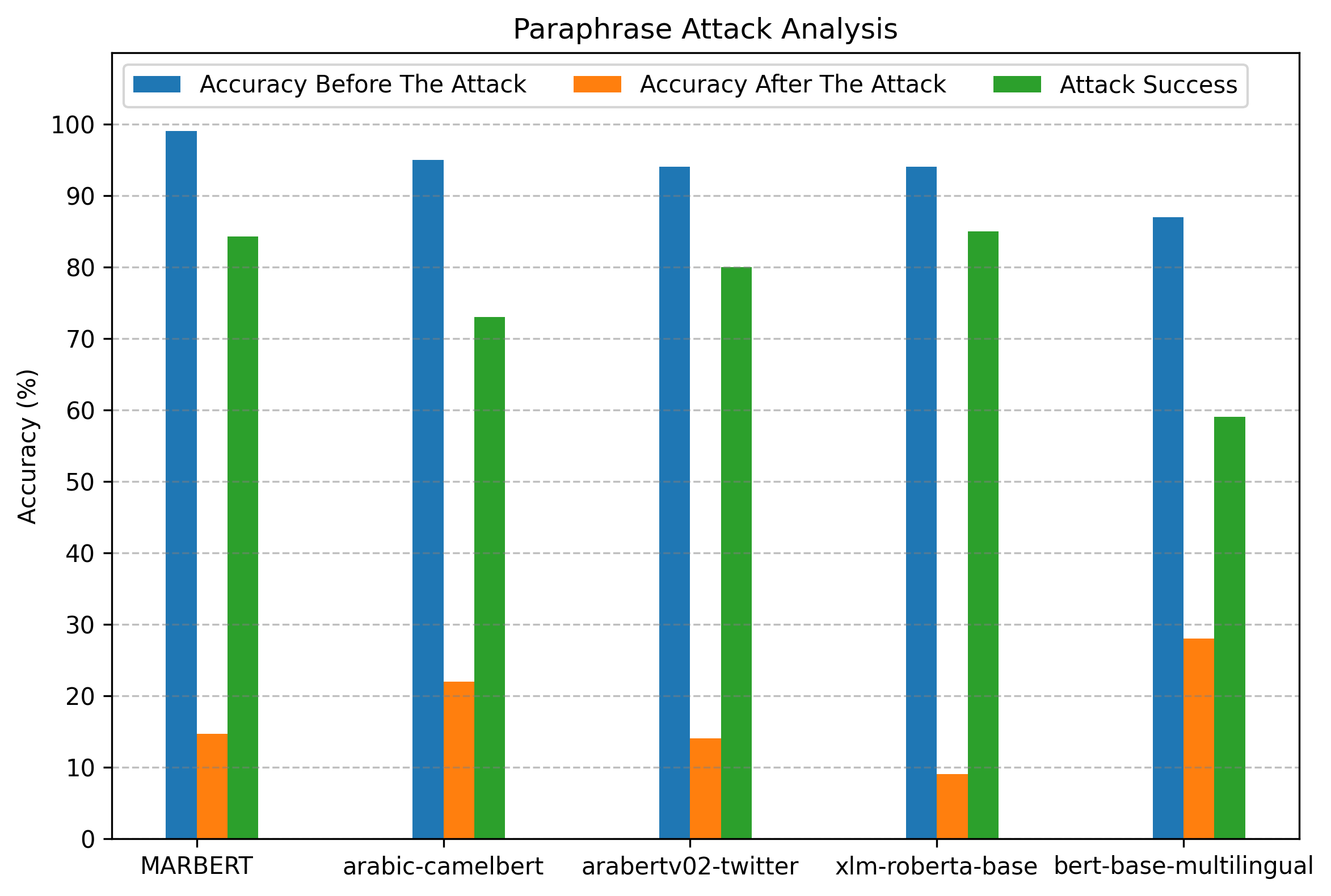}
\end{subfigure}
\caption{The results of the paraphrase-based attack on the victim models.}
\label{fig:Paraphrase}
\end{figure}

\begin{figure}[h!]
\begin{subfigure}{0.45\textwidth}
    \includegraphics[width=\linewidth]{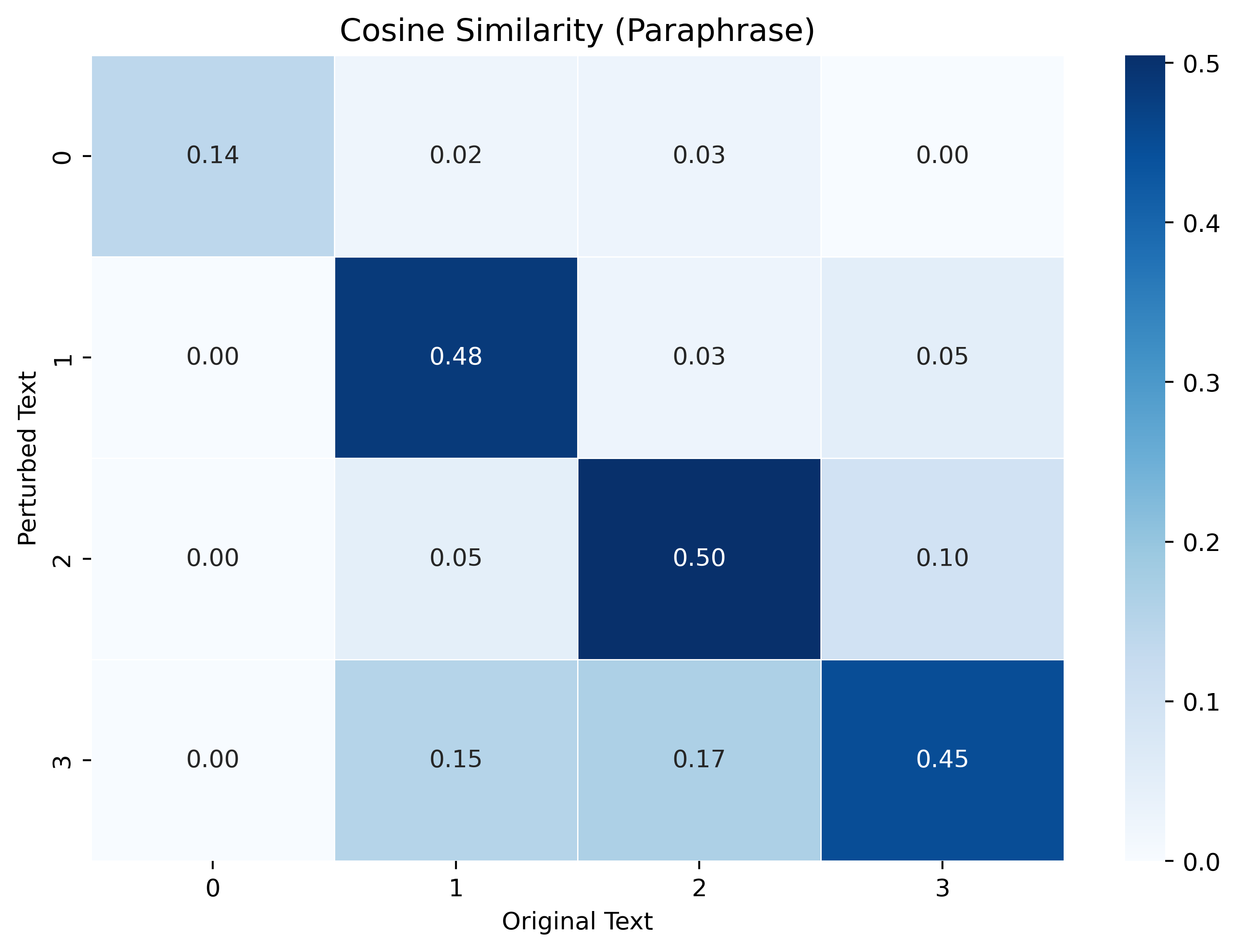} 
\end{subfigure}
\hfill 
\begin{subfigure}{0.45\textwidth}
    \includegraphics[width=\linewidth]{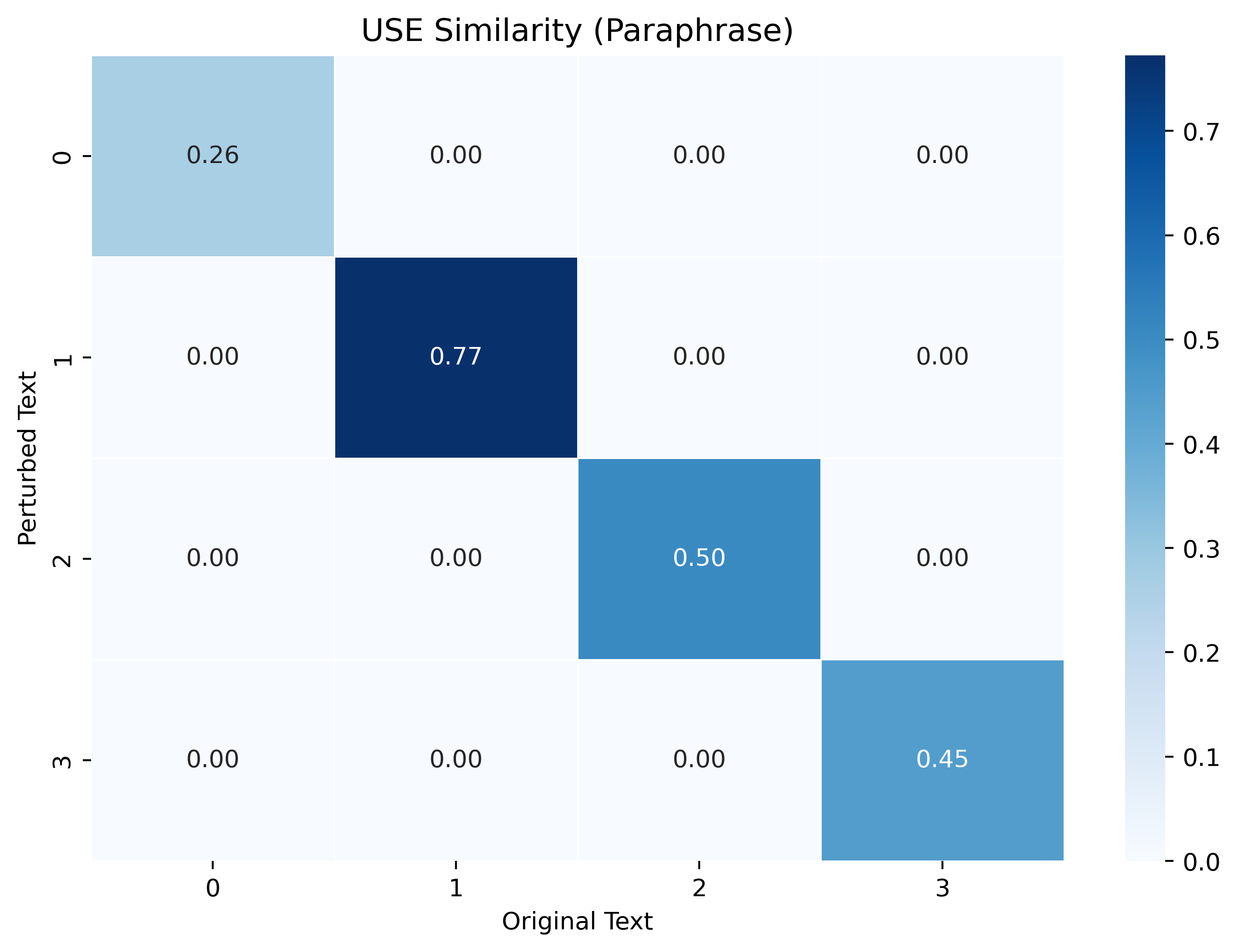}
\end{subfigure}
 \caption{Cosine and USE Similarity scores between original samples and their corresponding adversarial examples.}
\label{fig:Paraphrase-matrix}
\end{figure}

Unlike the previously discussed attacks, paraphrasing the sentences involves rephrasing the entire sequence while maintaining the original meaning. This method significantly alters the text's structure and wording but preserves its semantic content, making it a powerful technique for evaluating model robustness. It is worth noting that, unlike character-level and word-level attacks which generate model-specific adversarial examples tailored to each victim model, the paraphrase attack produces a single shared set of adversarial examples applied uniformly to all models. As a result, the perturbation and similarity metrics are identical across all victim models. The perturbation rate for paraphrasing attacks is notably high, reaching 83\%, indicating that a large portion of the text is changed during paraphrasing as shown in Figure \ref{fig:Paraphrase}. Despite this high perturbation rate, the paraphrased texts preserve semantic similarity, achieving an average Cosine Similarity of 39.43\% and a USE Similarity of 61.48\% across the test set, as illustrated in Figure \ref{fig:Paraphrase-matrix}. MARBERT is the most affected by an 85\% accuracy decrease, followed by XLM-T as shown in Table \ref{Attacks Results Paraphrase}.

In general, the five victim models were successfully attacked by our generated textual adversarial samples. All the models endured notable performance degradation, which shows the effectiveness of the attacks techniques. Moreover, the proposed methods have met the constraints of providing grammatically correct and semantically similar perturbed samples. Our experimental findings can be highlighted as follows:
\begin{itemize}
    \item Among the three character-level attacks, the diacritics attack had the heaviest impact on the models' performance as well as the lowest perturbation rates and highest similarity scores. 
    \item As for word-level attacks, manipulation of Arabic conjunctions maintained a low perturbation distance, more than 90\% Cosine Similarity, and a considerable accuracy reduction that goes up to 58\%.
    \item The paraphrase-based attack hindered the models' capability to correctly classify adversarial samples by an average of 76\% reduction in accuracy.
\end{itemize}
The above experimental results reveal critical weaknesses in the leading Arabic LMs' ability to handle punctuation insertions, visual and phonetic substitutions as well as syntactic variability. This underscores the need for developing more robust models that can accurately interpret and process diverse inputs without significant performance degradation.

\subsection{Adversarial Training}
As a defense mechanism, adversarial training \cite{goodfellow2014explaining,yuan2019adversarial} was conducted on the victim models. Although adversarial training induces a modest decline (roughly 1-2\%) in the clean accuracy of the fortified models, it substantially increased resilience. Another trade-off is the additional computational overhead, which increased the training time by approximately 25–40\%, which varies based on model size and attack complexity. 

Table~\ref{tab:char_word_attacks_merged} demonstrates the performance of five Arabic language models—AraBERT, MARBERT, CaMeLBERT, mBERT, and XLM-T on the distinct adversarial attacks after adversarial training. For each attack, the table shows the clean accuracy (pre-attack), the post-attack accuracy after adversarial training, the accuracy decrease, and the robustness improvement.

\begin{sidewaystable}
\centering
\small
\begin{tabular*}{\textwidth}{@{\extracolsep\fill}l l c c c c}
\toprule
\textbf{Model} & \textbf{Attack}   & \textbf{Pre‐attack Accuracy} & \textbf{Post‐attack Accuracy} & \textbf{Decrease} & \textbf{Improvement} \\
\midrule
\multirow{6}{*}{AraBERT}
    & Diacritic     & \multirow{6}{*}{94\%} & 22.61\% & 71.39\% & +20.61\% \\
    & Phonetic      &                       & \textbf{92.89\%} &  1.11\% & +1.89\%  \\
    & Visual        &                       & 20.87\% & \textbf{73.13\%} & +20.87\% \\
    & MLM           &                       & 87.16\% &  6.84\% & +6.16\%  \\
    & Conjunction   &                       & 49.33\% & 44.67\% & +13.33\% \\
    & Paraphrase    &                       & 52.30\% & 41.70\% & \textbf{+38.30\%} \\
\midrule
\multirow{6}{*}{MARBERT}
    & Diacritic     & \multirow{6}{*}{99\%} & 95.65\% &  3.35\% & +1.65\%  \\
    & Phonetic      &                       & 98.77\% &  0.23\% & +0.77\%  \\
    & Visual        &                       & 89.57\% &  9.43\% & +3.57\%  \\
    & MLM           &                       & 83.85\% & 15.15\% & +23.85\% \\
    & Conjunction   &                       & \textbf{98.89\% }&  0.11\% & +1.39\%  \\
    & Paraphrase    &                       & 62.40\% & \textbf{36.60\%} & \textbf{+48.40\%} \\
\midrule
\multirow{6}{*}{CaMeLBERT}
    & Diacritic     & \multirow{6}{*}{95\%} & 30.43\% & \textbf{64.57\%} & +15.43\% \\
    & Phonetic      &                       & 78.13\% & 16.87\% & +14.13\% \\
    & Visual        &                       & 33.91\% & 61.09\% & +16.91\% \\
    & MLM           &                       & \textbf{86.48\%} &  8.52\% & +15.78\% \\
    & Conjunction   &                       & 81.33\% & 13.67\% & +7.83\%  \\
    & Paraphrase    &                       & 67.30\% & 27.70\% & \textbf{+45.30\%} \\
\midrule
\multirow{6}{*}{mBERT}
    & Diacritic     & \multirow{6}{*}{87\%} & 55.65\% & \textbf{31.35\%} & +10.65\% \\
    & Phonetic      &                       & 80.43\% &  6.57\% & +15.43\% \\
    & Visual        &                       & 57.39\% & 29.61\% & +9.39\%  \\
    & MLM           &                       & 71.12\% & 15.88\% & +23.32\% \\
    & Conjunction   &                       & \textbf{85.78\% }&  1.22\% & +2.78\%  \\
    & Paraphrase    &                       & 60.60\% & 26.40\% & \textbf{+32.60\%} \\
\midrule
\multirow{6}{*}{XLM-T}
    & Diacritic     & \multirow{6}{*}{94\%} & 73.04\% & 20.96\% & +8.04\%  \\
    & Phonetic      &                       & \textbf{88.35\%} &  5.65\% & +8.35\%  \\
    & Visual        &                       & 79.13\% & 14.87\% & +6.13\%  \\
    & MLM           &                       & 78.33\% & 15.67\% & +28.33\% \\
    & Conjunction   &                       & 76.42\% & 17.58\% & +21.42\% \\
    & Paraphrase    &                       & 48.40\% & \textbf{45.60\%} & \textbf{+37.40\%} \\
\bottomrule
\end{tabular*}
\caption{Model performance after adversarial training.}
\label{tab:char_word_attacks_merged}
\end{sidewaystable}

Across all models, adversarial training led to consistent improvements in robustness. The notable findings of the adversarial training experiments are summarized below:

\begin{itemize}
  
    \item  Despite adversarial fine-tuning, Diacritic and Visual attacks remained highly effective in degrading performance. AraBERT dropped to 22.61\% (Diacritic) and 20.87\% (Visual), despite improvements over its pre-trained state. These results indicate that Diacritic and Visual attacks remain a challenge in most victim models robustness although adversarial training was conducted. This emphasizes the need for more targeted defense strategies.

    \item  The highest robustness gains were observed for word-level attacks, particularly MLM and the sentence-level attack Paraphrase. MARBERT and CaMeLBERT achieved improvements of +48.40\% and +45.30\% on Paraphrase attacks, respectively, which may indicate that their contextual embeddings are particularly effective at capturing underlying semantic meaning, enabling them to generalize better to such paraphrased inputs during adversarial training.

    \item \textbf{Model-Specific Behavior:}
    \begin{itemize}
        \item \textbf{AraBERT} exhibited substantial gains from adversarial training, especially on Diacritic, Visual, and Paraphrase attacks (over +20\% each).
        \item \textbf{MARBERT} remained the most consistently robust model overall, with minimal decreases across all attack types.
        \item \textbf{CaMeLBERT} demonstrated strong improvements in general, suggesting effective generalization to adversarial data.
        \item \textbf{mBERT} improved moderately but was less robust than Arabic-specific models.
        \item \textbf{XLM-T} performed well against word-level attacks, particularly with +37.40\% improvement on Paraphrase, highlighting the benefits of multilingual pre-training.
    \end{itemize}

\end{itemize}

 From the findings above, adversarial training enhanced robustness, but challenges persist with character-level noise in morphologically rich languages like Arabic. MARBERT proved most resilient overall, while AraBERT showed the greatest relative gains, emphasizing both the potential and limits of the conventional adversarial training approaches.

\section{Conclusion and Future Work}\label{sec5}

Adversarial attacks on Arabic language models (LMs) have ethical implications, particularly in critical applications such as healthcare, finance, and governance, where incorrect predictions can cause harm, bias, or misinformation. To address these concerns, researchers should explore and report model vulnerabilities through a proper adversarial robustness evaluation to achieve reliability. Hence, the objective of this paper was to study and evaluate the adversarial robustness of the most successful Arabic and multilingual NLP models against customized Arabic textual black-box attacks. Victim models were evaluated on a diverse set of adversarial examples at all levels of granularity, including three character-level attacks (diacritics, visual, and phonetic), two word-level attacks (MLM and conjunctions), and one sentence-level attack (paraphrase).  

The results reveal that AraBERT model is the weakest when it comes to diacrtics, visual, and conjunction attacks. It is followed by XLM-T, which endured the heaviest performance degradation after MLM attacks. In addition, CaMeLBERT showed the poorest post-attack performance when it was evaluated after dialectal manipulation and MARBERT had the most accuracy decrease when it was evaluated on paraphrased examples. Among the attacks, paraphrasing is the most effective in reducing model accuracy, leading to the highest average accuracy decrease. Diacritic manipulations prove most impactful for perturbation distance, while conjunction substitutions cause the most significant drop in similarity scores. Nevertheless, the use of adversarial training improved model robustness, particularly against higher-level word-based perturbations. However, resilience to character-level noise remains an open challenge. MARBERT stands out as the most robust overall, while AraBERT exhibits the largest relative improvements. These findings highlight both the effectiveness of adversarial attacks and the limitations of adversarial training, underscoring the need for further research to enhance models' ability to capture subtle character-level distortions.

This work can be extended through incorporating importance-based character and sentence perturbations to minimize the perturbation rate and increase the attack success and similarity scores. Attacks can also be used to evaluate the recent and popular GPT models similar to \cite{kadaoui2023tarjamat,khondaker2023gptaraeval} approaches. Furthermore, the models’ robustness can be evaluated on other NLP tasks such as dialect identification, named‐entity recognition, or question answering. Such cross‐task assessments would help identify any task‐specific vulnerabilities.
To enhance the robustness of Arabic language models (LMs), researchers and developers should strengthen model defenses through integrating customized defense techniques \cite{bountakas2023defense}, the information bottleneck method \cite{zhang2022improving}, and advanced frameworks like Interval Bound Propagation (IBP) \cite{gowal2018effectiveness,goyal2023survey}. 

Overall, this study presents the first comprehensive adversarial robustness evaluation targeting Arabic NLP models across multiple perturbation levels. Our findings contribute valuable insights into the practical vulnerabilities and defense limitations of widely used Arabic language models, setting a foundation for future work in secure and trustworthy Arabic NLP.

\subsection*{Funding}
 The authors received no specific
funding for this study.
\subsection*{Data availability}
The dataset \cite{Alajmi} that was used to train the models is available for public use on \href{https://huggingface.co/datasets/Anwaarma/MySentimentAnwarBig}{huggingface}.
\section*{Declarations}
\subsection*{Conflict of Interest}
The authors declare that they have no conflict of interest. 
\begin{appendices}

\section{SHAP Force Plots}\label{secA1}

\begin{figure}[H]
    \centering
    \includegraphics[width=\linewidth]{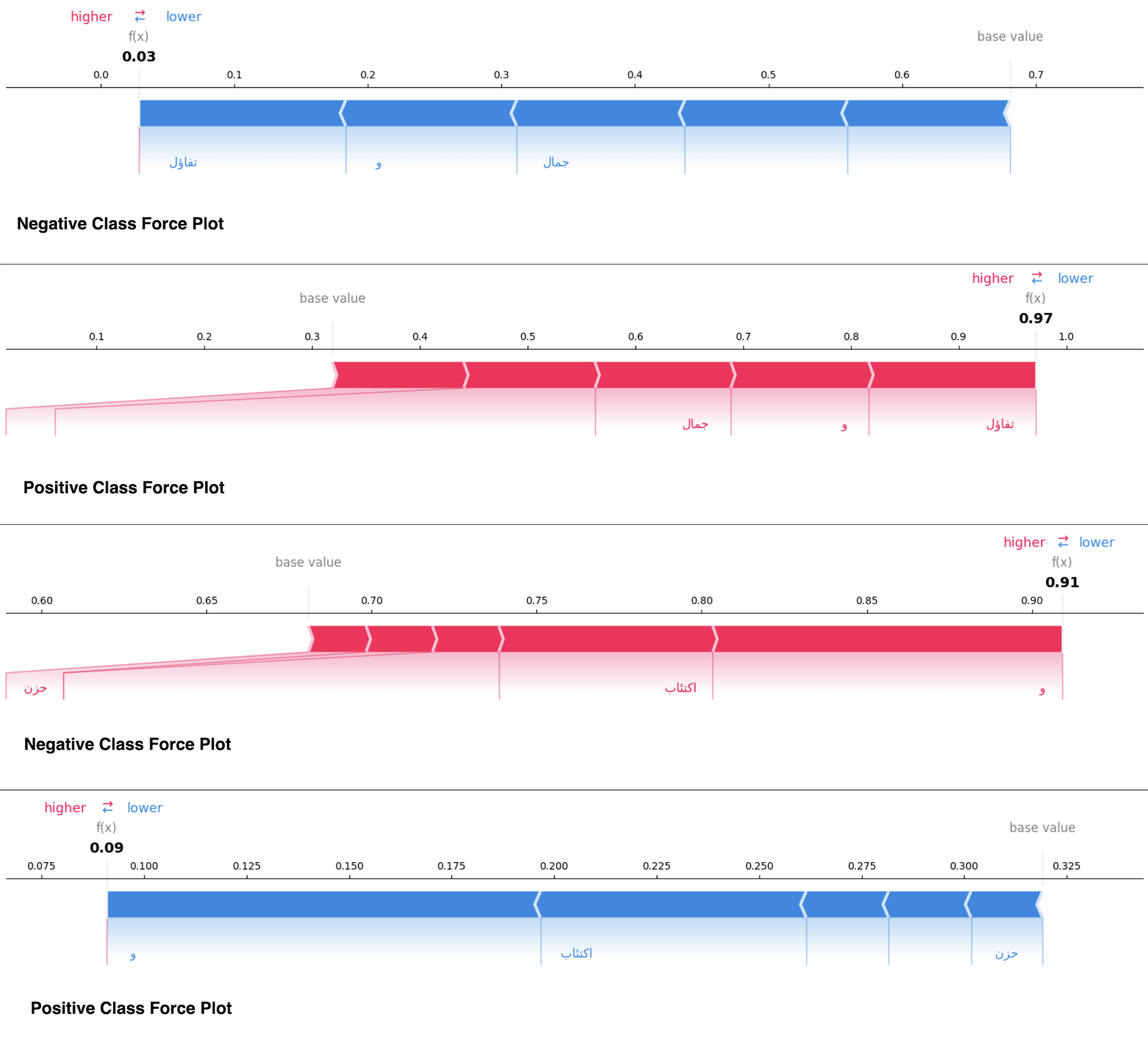}
    \caption{SHAP force plots for the first pair of examples.}
    \label{fig:a1}
\end{figure}

\begin{figure}[H]
    \centering
    \includegraphics[width=\linewidth]{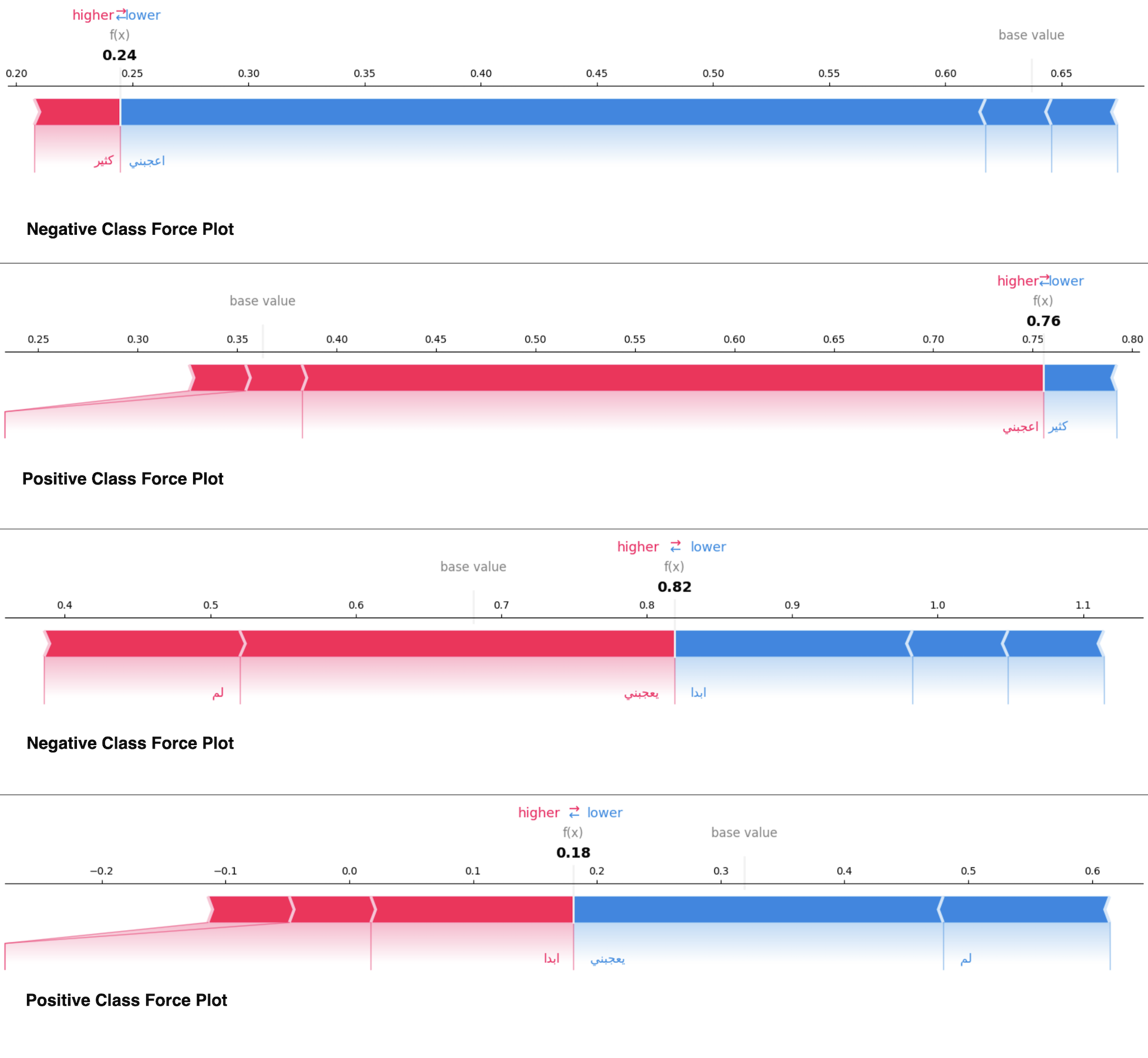}
    \caption{SHAP force plots for the second pair of examples.}
    \label{fig:a2}
\end{figure}

Figures~\ref{fig:a1} and \ref{fig:a2} in this appendix show SHAP force plots for four Arabic test sentences, generated using the AraBERT model fine-tuned for sentiment analysis. For each sentence, two subplots are presented: the first highlights token contributions driving the prediction toward the negative class, and the second highlights contributions toward the positive class. Horizontal bars indicate how each token shifts the model’s output from its expected value (base value) toward the selected class. Blue bars represent features that push the base value lower, whereas red bars represent features that push the prediction higher. 

\end{appendices}
\bibliography{sn-bibliography}

\end{document}